\documentclass[a4paper]{article}

\usepackage{amsmath,amsfonts,microtype,graphicx,array,hyperref,natbib,color,multirow,amsthm,subcaption,iclr2017_conference,times}

\iclrfinalcopy

\DeclareMathOperator*{\E}{\mathbb{E}}
\newtheorem{theorem}{Theorem}
\newtheorem{remark}{Remark}
\newcommand{\percent}{\%}

\title{Revisiting Classifier Two-Sample Tests}

\author{David Lopez-Paz$^1$, Maxime Oquab$^{1,2}$\\
$^1$Facebook AI Research, $^2$WILLOW project team, Inria / ENS / CNRS\\
\texttt{dlp@fb.com}, \texttt{maxime.oquab@inria.fr}
}

\begin{document}
  \maketitle

  \begin{abstract}
    The goal of two-sample tests is to assess whether two samples, $S_P \sim
    P^n$ and $S_Q \sim Q^m$, are drawn from the same distribution.  Perhaps
    intriguingly, one relatively unexplored method to build two-sample tests is
    the use of binary classifiers. In particular, construct a dataset by
    pairing the $n$ examples in $S_P$ with a positive label, and by pairing the
    $m$ examples in $S_Q$ with a negative label. If the null hypothesis ``$P =
    Q$'' is true, then the classification accuracy of a binary classifier on a
    held-out subset of this dataset should remain near chance-level.  As we
    will show, such \emph{Classifier Two-Sample Tests} (C2ST) learn a suitable
    representation of the data on the fly, return test statistics in interpretable 
    units, have a simple null distribution, and their predictive uncertainty
    allow to interpret where $P$ and $Q$ differ.

    The goal of this paper is to establish the properties, 
    performance, and uses of C2ST.  First, we analyze their main
    theoretical properties.  Second, we compare their performance against a
    variety of state-of-the-art alternatives.  Third, we propose their use to
    evaluate the sample quality of generative models with intractable
    likelihoods, such as Generative Adversarial Networks (GANs).  Fourth, we
    showcase the novel application of GANs together with C2ST for causal
    discovery.
  \end{abstract}

  \section{Introduction}
  One of the most fundamental problems in statistics is to assess whether two
  samples, $S_P \sim P^n$ and $S_Q \sim Q^m$, are drawn from the same
  probability distribution. To this end, \emph{two-sample tests}
  \citep{lehmann2006testing} summarize the differences between the two samples
  into a real-valued test \emph{statistic}, and then use the value of such
  statistic to accept\footnote{For clarity, we abuse statistical language and
  write ``accept'' to mean ``fail to reject''.} or reject the null hypothesis
  ``$P=Q$''. The development of powerful two-sample tests is instrumental in a
  myriad of applications, including the evaluation and comparison of generative
  models.  Over the last century, statisticians have nurtured a wide variety of
  two-sample tests. However, most of these tests are only applicable to
  one-dimensional examples, require the prescription of a fixed representation
  of the data, return test statistics in units that are difficult to interpret,
  or do not explain \emph{how} the two samples under comparison differ.

  Intriguingly, there exists a relatively unexplored strategy to build
  two-sample tests that overcome the aforementioned issues: training a binary
  classifier to distinguish between the examples in $S_P$ and the examples in
  $S_Q$.  Intuitively, if $P=Q$, the test accuracy of such binary classifier
  should remain near chance-level. Otherwise, if $P\neq Q$ and the binary
  classifier is able to unveil some of the distributional differences between
  $S_P$ and $S_Q$, its test accuracy should depart from chance-level.  As we
  will show, such \emph{Classifier Two-Sample Tests} (C2ST) learn a suitable
  representation of the data on the fly, return test statistics in interpretable 
  units, have simple asymptotic distributions, and their learned features and
  predictive uncertainty provide interpretation on \emph{how} $P$ and $Q$ differ.
  In such a way, this work brings together the communities of statistical testing
  and representation learning.
  
  The goal of this paper is to establish the theoretical properties and
  evaluate the practical uses of C2ST.  To this end, our \textbf{contributions}
  are:
  \begin{itemize}
    \item We review the basics of two-sample tests in Section~\ref{sec:two}, as
    well as their common applications to measure statistical dependence and
    evaluate generative models.
    \item We analyze the attractive properties of C2ST (Section~\ref{sec:neural})
    including an analysis of their exact asymptotic distributions, testing
    power, and interpretability.
    \item We evaluate C2ST on a wide variety of synthetic and real data
    (Section~\ref{sec:experiments}), and compare their performance against
    multiple state-of-the-art alternatives. Furthermore, we provide 
    examples to illustrate how C2ST can interpret the differences between pairs
    of samples.
    \item In Section~\ref{sec:gmodels}, we propose the use of classifier
    two-sample tests to evaluate the sample quality of generative models with
    intractable likelihoods, such as Generative Adversarial Networks
    \citep{goodfellow2014generative}, also known as GANs.
    \item As a novel application of the synergy between C2ST 
    and GANs, Section~\ref{sec:causality}
    proposes the use of these methods for causal discovery.
  \end{itemize}

  \section{Two-Sample Testing}\label{sec:two}
  The goal of two-sample tests is to assess whether two samples, denoted by
  $S_P \sim P^n$ and $S_Q \sim Q^m$, are drawn from the same distribution
  \citep{lehmann2006testing}.  More specifically, two-sample tests either
  accept or reject the \emph{null hypothesis}, often denoted by $H_0$, which
  stands for ``$P= Q$''. When rejecting $H_0$, we say that the two-sample test
  favors the \emph{alternative hypothesis}, often denoted by $H_1$, which
  stands for ``$P \neq Q$''. To accept or reject $H_0$, two-sample tests
  summarize the differences between the two \emph{samples} (sets of identically
  and independently distributed \emph{examples}):
  \begin{align}
    S_P := \{ x_1, \ldots, x_n \} \sim P^n(X) \text{ and } S_Q := \{ y_1,
    \ldots, y_m \} \sim Q^m(Y)\label{eq:samples}
  \end{align}
  into a statistic $\hat{t} \in \mathbb{R}$.  Without loss of generality, we
  assume that the two-sample test returns a small statistic when
  the {null hypothesis} ``$P=Q$'' is true, and a large statistic otherwise.
  Then, for a sufficiently small statistic, the two-sample test will accept
  $H_0$.  Conversely, for a sufficiently large statistic, the two-sample test
  will reject $H_0$ in favour of $H_1$.

  More formally, the statistician performs a two-sample test in four steps.
  First, decide a \emph{significance level} $\alpha \in [0,1]$, which is an
  input to the two-sample test. Second, compute the two-sample test statistic
  $\hat{t}$.  Third, compute the \emph{p-value} $\hat{p} = P(T \geq \hat{t} |
  H_0)$, the probability of the two-sample test returning a statistic as large
  as $\hat{t}$ when $H_0$ is true. Fourth, reject $H_0$ if $\hat{p} < \alpha$,
  and accept it otherwise. 

  Inevitably, two-sample tests can fail in two different ways.  First, to make
  a \emph{type-I error} is to reject the null hypothesis when it is true (a
  ``false positive''). By the definition of $p$-value, the probability of making a type-I error is
  upper-bounded by the significance level $\alpha$.  Second, to make a
  \emph{type-II error} is to accept the null hypothesis when it is false (a
  ``false negative'').  We denote the probability of making a type-II error by
  $\beta$, and refer to the quantity $\pi = 1-\beta$ as the \emph{power} of
  a test. Usually, the statistician uses domain-specific knowledge to evaluate
  the consequences of a type-I error, and thus prescribe an appropriate
  significance level $\alpha$.  Within the prescribed significance level
  $\alpha$, the statistician prefers the two-sample test with maximum power
  $\pi$.

  Among others, two-sample tests serve two other
  uses.
  First, two-sample tests can \emph{measure statistical dependence}
  \citep{gretton2012kernel}.  In particular, testing the independence null
  hypothesis ``the random variables $X$ and $Y$ are independent'' is 
  testing the two-sample null hypothesis ``$P(X,Y)=P(X)P(Y)$''. In
  practice, the two-sample test would compare the sample $S =
  \{(x_i,y_i)\}_{i=1}^n \sim P(X,Y)^n$ to a sample $S_\sigma = \{(x_i,
  y_{\sigma(i)})\}_{i=1}^n \sim (P(X)P(Y))^n$, where $\sigma$ is a random
  permutation of the set of indices $\{1, \ldots, n \}$.
  This approach is consistent when considering all possible random
  permutations. However, since independence testing is a subset of two-sample
  testing, specialized independence tests may exhibit higher power for this
  task \citep{gretton2005measuring}.

  Second, two-sample tests can \emph{evaluate the sample quality of
  generative models} with intractable likelihoods, but tractable sampling
  procedures.  Intuitively, a generative model produces good samples $\hat{S} =
  \{\hat{x}_i\}_{i=1}^n$ if these are indistinguishable from the real data
  $S=\{x_i\}_{i=1}^n$ that they model.  Thus, the two-sample test statistic
  between $\hat{S}$ and $S$ measures the fidelity of the samples $\hat{S}$
  produced by the generative model. The use of two-sample tests to evaluate the
  sample quality of generative models include the pioneering work of
  \citet{box1980sampling}, the use of Maximum Mean Discrepancy (MMD) criterion
  \citep{bengio2013bounding,karolina,lloyd2015statistical,bounliphone2015test,gretton2016},
  and the connections to density-ratio estimation
  \citep{kanamori2010,Wornowizki2016,Menon2016,Mohamed2016}.

  Over the last century, statisticians have nurtured a wide variety of
  two-sample tests. Classical two-sample tests include the $t$-test
  \citep{student1908probable}, which tests for the difference in means of two
  samples; the Wilcoxon-Mann-Whitney test
  \citep{wilcoxon1945individual,mann1947test}, which tests for the difference
  in rank means of two samples; and the Kolmogorov-Smirnov tests
  \citep{kolmogorov1933sulla,smirnov1939estimation} and their variants
  \citep{kuiper}, which test for the difference in the empirical cumulative
  distributions of two samples. However, these classical tests are only
  efficient when applied to one-dimensional data. Recently, the use of kernel
  methods \citep{smola98} enabled the development of two-sample tests
  applicable to multidimensional data. Examples of these tests include the MMD
  test \citep{gretton2012kernel}, which looks for differences in the empirical
  kernel mean embeddings of two samples, and the Mean Embedding test or ME
  \citep{jit2,metests}, which looks for differences in the empirical kernel
  mean embeddings of two samples at optimized locations. However, kernel
  two-sample tests require the prescription of a manually-engineered
  representation of the data under study, and return values in units that are
  difficult to interpret.  Finally, only the ME test provides a mechanism to
  interpret how $P$ and $Q$ differ.

  Next, we discuss a simple but relatively unexplored strategy to build two-sample tests
  that overcome these issues: the use of binary classifiers.

  \section{Classifier Two-Sample Tests (C2ST)}\label{sec:neural}

  Without loss of generality, we assume access to the two samples $S_P$ and
  $S_Q$ defined in \eqref{eq:samples}, where $x_i, y_j \in \mathcal{X}$, for
  all $i = 1, \ldots, n$ and $j = 1, \ldots, m$, and $m=n$. To test whether the null
  hypothesis $H_0 : P=Q$ is true, we proceed in five steps.  First, construct
  the dataset
  \begin{equation*}
    \mathcal{D} = \{(x_i, 0)\}_{i=1}^n \cup \{(y_i, 1)\}_{i=1}^n =: \{(z_i, l_i)\}_{i=1}^{2n}.
  \end{equation*}
  Second, shuffle $\mathcal{D}$ at random, and split it into the disjoint \emph{training} and {testing}
  subsets $\mathcal{D}_\text{tr}$ and $\mathcal{D}_\text{te}$, where 
  $\mathcal{D} = \mathcal{D}_\text{tr} \cup \mathcal{D}_\text{te}$ and
  $n_\text{te} := |\mathcal{D}_\text{te}|$. Third, train a binary classifier $f
  : \mathcal{X} \to [0,1]$ on $\mathcal{D}_\text{tr}$; in the following, we assume
  that $f(z_i)$ is an estimate of the conditional probability distribution
  $p(l_i = 1 | z_i)$. Fourth, return the 
  classification accuracy on $\mathcal{D}_\text{te}$:
  \begin{equation}\label{eq:stat}
    \hat{t} = \frac{1}{n_\text{te}} \sum_{(z_i,l_i) \in \mathcal{D}_\text{te}}
    \mathbb{I}\left[ \mathbb{I}\left(f(z_i) > \frac{1}{2}\right) = l_i \right]
  \end{equation}
  as our \emph{C2ST statistic}, where $\mathbb{I}$ is the
  indicator function. The intuition here is that if $P=Q$, the test accuracy
  \eqref{eq:stat} should remain near
  chance-level.  In opposition, if $P \neq Q$ and the binary classifier unveils
  distributional differences between the two samples, the test classification
  accuracy \eqref{eq:stat} should be \emph{greater} than chance-level.
  Fifth, to accept or reject the null hypothesis, compute a
  p-value using the null distribution of the C2ST, as
  discussed next.

  \subsection{Null and Alternative Distributions}\label{sec:null}
  Each term $\mathbb{I}\left[ \mathbb{I}(f(z_i) > 1/2) = l_i\right]$ appearing
  in \eqref{eq:stat} is an independent $\text{Bernoulli}(p_i)$ random variable,
  where $p_i$ is the probability of classifying correctly the example $z_i$ in
  $\mathcal{D}_\text{te}$.
  
  First, under the null hypothesis $H_0 : P = Q$, the samples $S_P \sim P^n$
  and $S_Q \sim Q^m$ follow the same distribution, leading to an impossible
  binary classification problem. In that case, $n_\text{te} \hat{t}$ follows a
  $\text{Binomial}(n_\text{te},p=\frac{1}{2})$ distribution. Therefore, for large
  $n_{\text{te}}$, we can use the central limit theorem to approximate the
  null distribution of \eqref{eq:stat} by $\mathcal{N}(\frac{1}{2}, \frac{1}{4 n_\text{te}})$.
  
  Second, under the alternative hypothesis $H_1 : P \neq Q$, the statistic
  $n_{\text{te}} \hat{t}$ follows a Poisson Binomial distribution, since the
  constituent Bernoulli random variables may not be identically distributed.
  In the following, we will approximate such Poisson Binomial distribution by
  the $\text{Binomial}(n,\bar{p})$ distribution, where $\bar{p} = \frac{1}{n}
  \sum_{i=1}^n p_i$ \citep{Ehm91}.
  Therefore, we can use the central limit theorem to approximate the
  alternative distribution of \eqref{eq:stat} by $\mathcal{N}(\bar{p},
  \frac{\bar{p}(1-\bar{p})}{n_\text{te}})$.
  
  \subsection{Testing power}
  To analyze the power (probability of correctly rejecting false null
  hypothesis) of C2ST, we assume that the our
  classifier has an expected (unknown) accuracy of $H_0: t = \frac{1}{2}$ under the null
  hypothesis ``$P=Q$'', and an expected accuracy of $H_1: t = \frac{1}{2} +
  \epsilon$ under the alternative hypothesis ``$P\neq Q$'', where $\epsilon \in
  (0, \frac{1}{2})$ is the \emph{effect size} distinguishing $P$ from $Q$. Let
  $\Phi$ be the Normal cdf, $n_\text{te}$ the number of samples available for testing,
  and $\alpha$ the significance level. Then,

  \begin{theorem}\label{thm:thm1}
    Given the conditions described in the previous paragraph, the approximate power of 
    the statistic \eqref{eq:stat} is $\Phi\left(
    \frac{\epsilon\sqrt{n_\text{te}}-\Phi^{-1}(1-\alpha)/2}{\sqrt{\frac{1}{4}-\epsilon^2}}\right)$.
  \end{theorem}
  See Appendix~\ref{app:proof} for a proof. The power bound in
  Theorem~\ref{thm:thm1} has an optimal order of magnitude for
  multi-dimensional problems
  \citep{bai1996effect,gretton2012kernel,reddi2015high}. These are problems
  with fixed $d$ and $n \to \infty$, where the power bounds do not depend on
  $d$.

  \begin{remark}
  We leave for future work the study of quadratic-time C2ST with optimal power
  in high-dimensional problems \citep{Ramdas15}. These are problems where the
  ratio $n/d \to c \in [0,1]$, and the power bounds
  depend on $d$. One possible line of research in this direction is to
  investigate the power and asymptotic distributions of quadratic-time
  C2ST statistics $
  \frac{1}{n_\text{te}(n_\text{te}-1)}\sum_{i\neq j} \mathbb{I}[
  \mathbb{I}(f(z_i,z_j)>\frac{1}{2}) = l_i]$, where the classifier $f(z,z')$ 
  predicts if the examples $(z,z')$ come from the same
  sample.
  \end{remark}

  Theorem~\ref{thm:thm1} also illustrates that maximizing the power of a
  C2ST is a trade-off between two competing objectives:
  choosing a classifier that \emph{maximizes the test accuracy} $\epsilon$ and
  \emph{maximizing the size of the test set} $n_{\text{te}}$. This relates 
  to the well known bias-variance trade-off in machine learning.
  Indeed, simple classifiers will miss more nonlinear patterns in the
  data (leading to smaller test accuracy), but call for less training data
  (leading to larger test set sizes).
  On the other hand, flexible classifiers will miss less nonlinear patterns in
  the data (leading to higher test accuracy), but call for more training data
  (leading to smaller test sizes). Formally, the relationship between the test
  accuracy, sample size, and the flexibility of a classifier depends on  
  capacity measures such as the VC-Dimension
  \citep{vapnik98}. Note that there is no restriction to perform model
  selection (such as cross-validation) on $\mathcal{D}_\text{tr}$.

  \begin{remark}\label{remark:loss}
  We have focused on test statistics \eqref{eq:stat} built on top of the
  zero-one loss $\ell_{0-1}(y,y') = \mathbb{I}[y=y'] \in \{0,1\}$. These
  statistics give rise to Bernoulli random variables, which can exhibit high
  variance. However, our arguments are readily extended to real-valued binary
  classification losses. Then, the variance of such real-valued losses would
  describe the norm of the decision function of the classifier two-sample test,
  appear in the power expression from Theorem~\ref{thm:thm1}, and serve as a
  hyper-parameter to maximize power as in \citep[Section
  3]{gretton2012optimal}.\footnote{For a related discussion on this issue, we
  recommend the insightful comment by Arthur Gretton and Wittawat Jitkrittum,
  available at \url{https://openreview.net/forum?id=SJkXfE5xx}.}
  \end{remark}

  \subsection{Interpretability}

  There are three ways to interpret the result of a C2ST. First, recall that
  the classifier predictions $f(z_i)$ are estimates of the conditional
  probabilities $p(l_i = 1 |z_i)$ for each of the samples $z_i$ in the test
  set. Inspecting these probabilities together with the true labels $l_i$
  determines which examples were correctly or wrongly labeled by the classifier,
  with the least or the most confidence. Therefore, the values $f(z_i)$ explain
  \emph{where} the two distributions differ.  Second, C2ST inherit the
  interpretability of their classifiers to explain which \emph{features} are
  most important to distinguish distributions, in the same way as the ME
  test \citep{metests}.  Examples of interpretable features include the filters
  of the first layer of a neural network, the feature importance of random
  forests, the weights of a generalized linear model, and so on.  Third, C2ST
  return statistics $\hat{t}$ in interpretable units: these
  relate to the percentage of samples correctly distinguishable between the two
  distributions. These interpretable numbers can complement the use of
  $p$-values. 

  \subsection{Prior Uses}

  The reduction of two-sample testing to binary classification was introduced
  in \citep{friedman2003multivariate}, studied within the context of
  information theory in
  \citep{perez2009estimation,reid2011information}, discussed in
  \citep{Fukumizu2009,gretton2012kernel}, and analyzed (for the case of linear
  discriminant analysis) in \citep{ramdas16}. The use of binary classifiers for
  two-sample testing is increasingly common in neuroscience: see
  \citep{pereira2009machine,olivetti2012induction} and the references therein.
  Implicitly, binary classifiers also perform two-sample tests in algorithms
  that discriminate data from noise, such as unsupervised-as-supervised
  learning \citep{friedman2001elements}, noise contrastive estimation
  \citep{gutmann2012noise}, negative sampling \citep{mikolov2013distributed},
  and GANs \citep{goodfellow2014generative}.

  \section{Experiments on Two-Sample Testing}\label{sec:experiments}

  We study two variants of classifier-based two-sample tests
  (C2ST): one based on neural networks (C2ST-NN), and one based
  on $k$-nearest neighbours (C2ST-KNN).  C2ST-NN has one
  hidden layer of 20 ReLU neurons, and trains for $100$ epochs using the Adam
  optimizer \citep{adam}. C2ST-KNN
  uses $k = \lfloor n_{\text{tr}}^{1/2}\rfloor$ nearest neighbours for
  classification.
  Throughout our experiments, we did not observe a significant improvement in
  performance when increasing the flexibility of these classifiers (e.g.,
  increasing the number of hidden neurons or decreasing the number of nearest
  neighbors).
  When analyzing one-dimensional data, we compare the performance of C2ST-NN
  and C2ST-KNN against the Wilcoxon-Mann-Whitney test
  \citep{wilcoxon1945individual,mann1947test}, the Kolmogorov-Smirnov test
  \citep{kolmogorov1933sulla,smirnov1939estimation}, and the Kuiper test
  \citep{kuiper}. In all cases, we also compare the performance of C2ST-NN and
  C2ST-KNN against the linear-time estimate of the Maximum Mean Discrepancy
  (MMD) criterion \citep{gretton2012kernel}, the ME test \citep{metests}, and
  the SCF test \citep{metests}.  We use a significance level $\alpha =0.05$
  across all experiments and tests, unless stated otherwise. We use Gaussian
  approximations to compute the null distributions of C2ST-NN and C2ST-KNN.  We
  use the implementations of the MMD, ME, and SCF tests gracefully provided by
  \citet{metests}, the scikit-learn implementation of the Kolmogorov-Smirnov
  and Wilcoxon tests, and the implementation from
  {\small\url{https://github.com/aarchiba/kuiper}} of the Kuiper test. The
  implementation of our experiments is available at
  {\small\url{https://github.com/lopezpaz/classifier_tests}}.

  \begin{figure}
    \begin{center}
      \includegraphics[width=\textwidth]{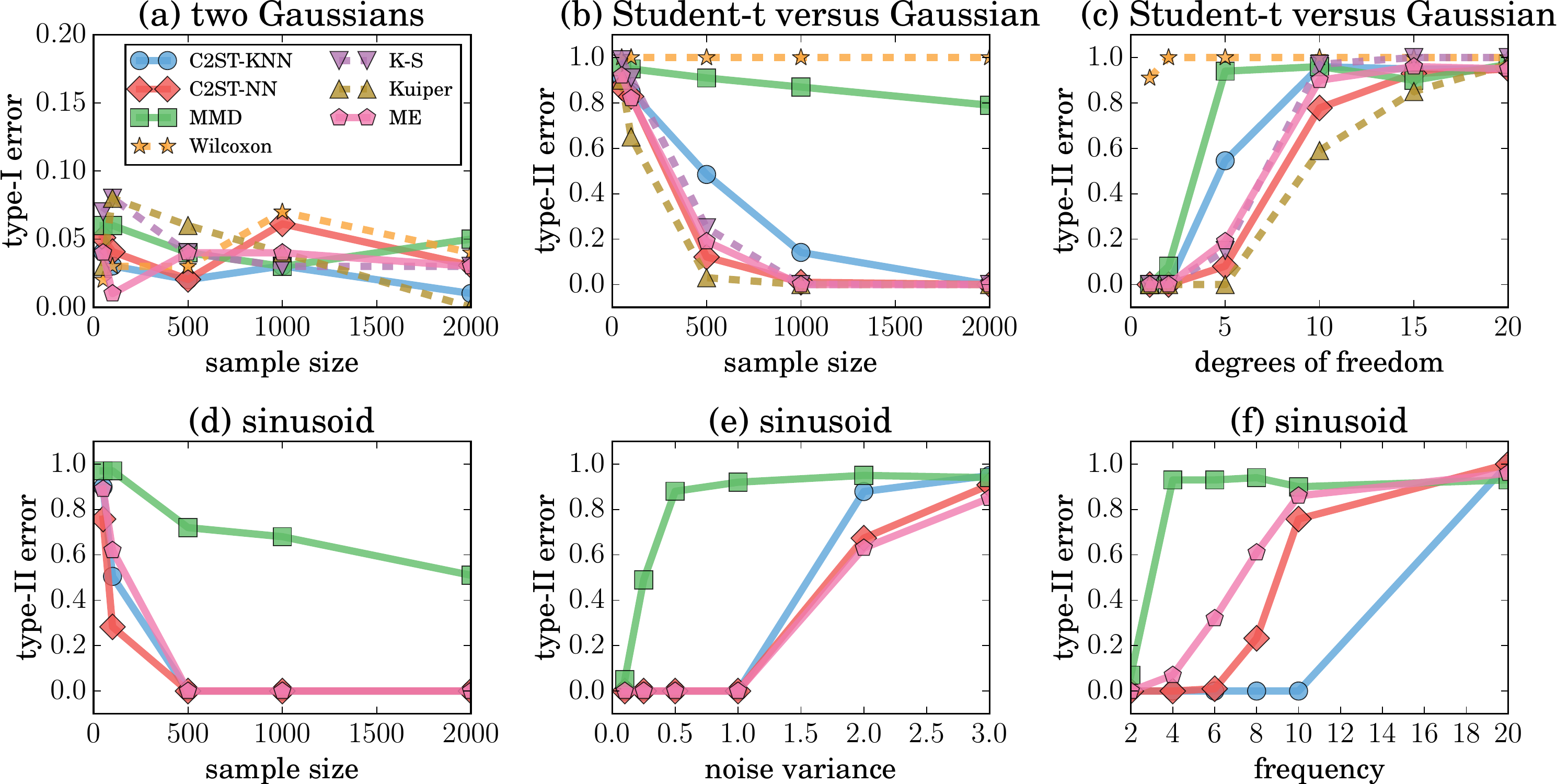}
    \end{center}
    \vspace{-0.2cm}
    \caption{Results (type-I and type-II errors) of our synthetic two-sample test experiments.}
    \vspace{-0.2cm}
    \label{fig:synthetic}
  \end{figure}

  \subsection{Experiments on Two-Sample Testing}\label{sec:smallexps}

  \paragraph{Control of type-I errors} We start by evaluating the correctness of
  all the considered two-sample tests by examining if the prescribed
  significance level $\alpha = 0.05$ upper-bounds their type-I error. To do so,
  we draw $x_1, \ldots, x_n, y_1, \ldots, y_n \sim \mathcal{N}(0,1)$, and run
  each two-sample test on the two samples $\{x_i\}_{i=1}^n$ and
  $\{y_i\}_{i=1}^n$. In this setup, a type-I error would be to reject the true
  null hypothesis.  Figure~\ref{fig:synthetic}(a) shows that the type-I error
  of all tests is upper-bounded by the prescribed significance level, for all
  $n\in \{25,50,100,500,1000,5000,10000\}$ and $100$ random repetitions. Thus,
  all tests control their type-I error as expected, up to random variations due
  to finite experiments.

  \paragraph{Gaussian versus Student} We consider distinguishing between
  samples drawn from a Normal distribution and samples drawn from a Student's
  t-distribution with $\nu$ degrees of freedom.  We shift and scale both samples
  to exhibit zero-mean and unit-variance. Since the Student's t distribution
  approaches the Normal distribution as $\nu$ increases, a two-sample test must
  focus on the peaks of the distributions to
  distinguish one from another.  Figure~\ref{fig:synthetic}(b,c) shows the percentage of type-II errors 
  made by all tests as we vary separately $n$ and $\nu$, over $100$
  trials (random samples). We set $n = 2000$ when $\nu$ varies, and let $\nu =
  3$ when $n$ varies. The Wilcoxon-Mann-Whitney exhibits the worst performance,
  as expected (since the ranks mean of the Gaussian and Student's t
  distributions coincide) in this experiment. The best performing method is the
  the one-dimensional Kuiper test, followed closely by the multi-dimensional
  tests C2ST-NN and ME.

  \paragraph{Independence testing on sinusoids} For completeness, we showcase
  the use two-sample tests to measure statistical dependence. This can be done,
  as described in Section~\ref{sec:two}, by performing a two-sample test
  between the observed data
  $\{(x_i,y_i)\}_{i=1}^n$ and $\{(x_i,y_{\sigma(i)})\}_{i=1}^n$, where $\sigma$
  is a random permutation. Since the distributions $P(X)P(Y)$ and $P(X,Y)$ are
  bivariate, only the C2ST-NN,
  C2ST-KNN, MMD, and ME tests compete in this task.  We draw $(x_i,
  y_i)$ according to the generative model $x_i \sim \mathcal{N}(0,1)$,
  $\epsilon_i \sim \mathcal{N}(0,\gamma^2)$, and $y_i \sim \cos(\delta x_i) +
  \epsilon_i$.  Here, $x_i$ are iid examples from the random variable $X$, and
  $y_i$ are iid examples from the random variable $Y$. Thus, the statistical
  dependence between $X$ and $Y$ weakens as we increase the frequency $\delta$ of
  the sinusoid, or increase the variance $\gamma^2$ of the additive noise.
  Figure~\ref{fig:synthetic}(d,e,f) shows the percentage of type-II errors made by C2ST-NN,
  C2ST-KNN, MMD, and ME as we vary separately $n$, $\delta$, and $\gamma$ over $100$ trials.
  We let $n = 2000$, $\delta = 1$, $\gamma = 0.25$ when fixed.
  Figure~\ref{fig:synthetic}(d,e,f) reveals that among all tests, C2ST-NN is
  the most efficient in terms of sample size, C2ST-KNN is the most robust with
  respect to high-frequency variations, and that C2ST-NN and ME are the most
  robust with respect to additive noise.

  \paragraph{Distinguishing between NIPS articles} We consider the problem of
  distinguishing between some of the categories of the 5903 articles published in
  the Neural Information Processing Systems (NIPS) conference from 1988 to
  2015, as discussed in \citet{metests}.
  We consider articles on Bayesian inference (Bayes), neuroscience (Neuro),
  deep learning (Deep), and statistical learning theory (Learn).
  Table~\ref{table:nips} shows the type-I errors (Bayes-Bayes row) and
  powers (rest of rows) for the tests reported in
  \citep{metests}, together with C2ST-NN, at a significance level $\alpha =
  0.01$, when averaged over $500$ trials. In these experiments, C2ST-NN
  achieves maximum power, while upper-bounding its type-I error by $\alpha$.

  \paragraph{Distinguishing between facial expressions} Finally, we apply
  C2ST-NN to the problem of distinguishing between positive (happy,
  neutral, surprised) and negative (afraid, angry, disgusted) facial
  expressions from the Karolinska Directed Emotional Faces dataset, as
  discussed in \citep{metests}. See the fourth plot of
  Figure~\ref{fig:interpretability}, first two-rows, for one example of each of
  these six emotions. Table~\ref{table:expressions} shows the type-I errors
  ($\pm$ vs $\pm$ row) and the powers ($+$ vs $-$ row) 
  for the tests reported in \citep{metests}, together with C2ST-NN, at $\alpha = 0.01$, 
  averaged over $500$ trials. C2ST-NN achieves a near-optimal
  power, only marginally behind the perfect results of SCF-full and
  MMD-quad.

  \begin{table}
  \resizebox{\textwidth}{!}{
  \begin{tabular}{|l|c|c|c|c|c|c|c|c|}
  \hline
  Problem & $n^{te}$ & ME-full & ME-grid & SCF-full & SCF-grid & MMD-quad & MMD-lin & C2ST-NN\\
  \hline
  Bayes-Bayes & 215 & .012 & .018 & .012 & .004 & .022 & .008 & \textbf{.002} \\ \hline\hline
  Bayes-Deep  & 216 & .954 & .034 & .688 & .180 & .906 & .262 & \textbf{1.00} \\ \hline
  Bayes-Learn & 138 & .990 & .774 & .836 & .534 & \textbf{1.00} & .238 & \textbf{1.00} \\ \hline
  Bayes-Neuro & 394 & \textbf{1.00} & .300 & .828 & .500 & .952 & .972 & \textbf{1.00} \\ \hline
  Learn-Deep  & 149 & .956 & .052 & .656 & .138 & .876 & .500 & \textbf{1.00} \\ \hline
  Learn-Neuro & 146 & .960 & .572 & .590 & .360 & \textbf{1.00} & .538 & \textbf{1.00} \\
  \hline
  \end{tabular}
  }
  \vspace{-0.2cm}
  \caption{Type-I errors (first row) and powers (rest of rows) in distinguishing
  NIPS papers categories.}
  \vspace{-0.2cm}
  \label{table:nips}
  \end{table}

  \begin{table}
  \resizebox{\textwidth}{!}{
  \begin{tabular}{|l|c|c|c|c|c|c|c|c|}
  \hline
  Problem & $n^{te}$ & ME-full & ME-grid & SCF-full & SCF-grid & MMD-quad & MMD-lin & C2ST-NN \\
  \hline
  $\pm$ vs. $\pm$ & 201 & .010 & .012 & .014 & \textbf{.002} & .018 & .008 & \textbf{.002} \\ \hline\hline
  $+$   vs. $-$   & 201 & {.998} & .656 & \textbf{1.00} & .750 & \textbf{1.00} & .578 & {.997} \\
  \hline
  \end{tabular}\\
  }
  \vspace{-0.2cm}
  \caption{Type-I errors (first row) and powers (second row) in distinguishing 
  facial expressions.}
  \vspace{-0.2cm}
  \label{table:expressions}
  \end{table}

  \section{Experiments on Generative Adversarial Network Evaluation}\label{sec:gmodels}
  
  \begin{table}
  \resizebox{\textwidth}{!}{
  \begin{tabular}{cccc}
      random sample & MMD & KNN & NN \\\hline
      \includegraphics[width=\textwidth,trim={1536px 0 0 0},clip]{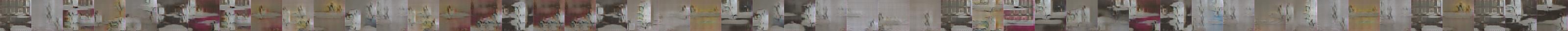}   & 0.158 & 0.830 & 0.999\\
      \includegraphics[width=\textwidth,trim={1536px 0 0 0},clip]{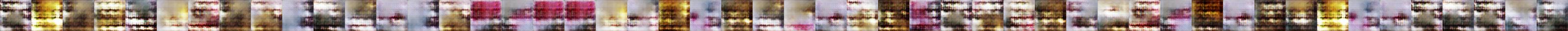}    & 0.154 & 0.994 & 1.000\\
      \includegraphics[width=\textwidth,trim={1536px 0 0 0},clip]{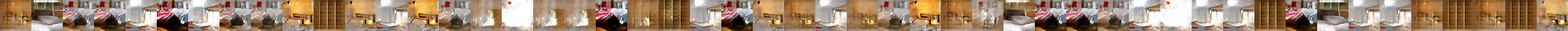}  & 0.048 & 0.962 & 1.000\\ \hline
      \includegraphics[width=\textwidth,trim={1536px 0 0 0},clip]{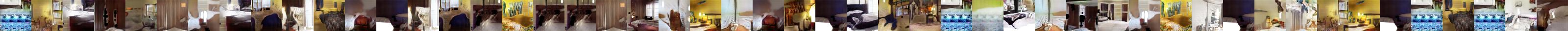}  & \underline{\bf 0.012} & 0.798 & 0.964\\
      \includegraphics[width=\textwidth,trim={1536px 0 0 0},clip]{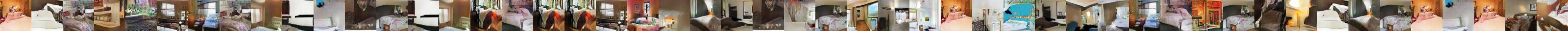}   & 0.024 & 0.748 & \underline{\bf 0.949}\\
      \includegraphics[width=\textwidth,trim={1536px 0 0 0},clip]{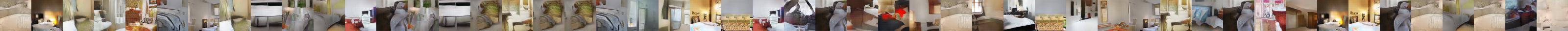}  & 0.019 & \underline{\bf 0.670} & 0.983\\\hline
      \includegraphics[width=\textwidth,trim={1536px 0 0 0},clip]{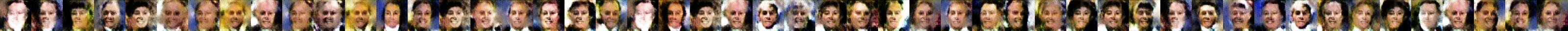}   & 0.152 & 0.940 & 1.000\\
      \includegraphics[width=\textwidth,trim={1536px 0 0 0},clip]{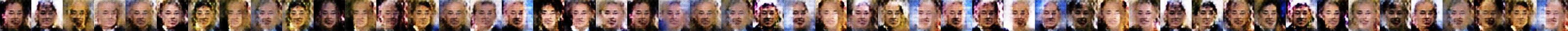}   & 0.222 & 0.978 & 1.000\\
      \includegraphics[width=\textwidth,trim={1536px 0 0 0},clip]{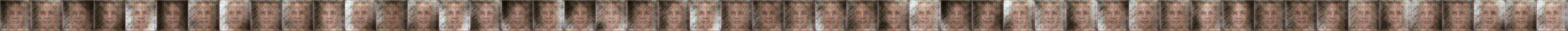}    & 0.715 & 1.000 & 1.000\\\hline
      \includegraphics[width=\textwidth,trim={1536px 0 0 0},clip]{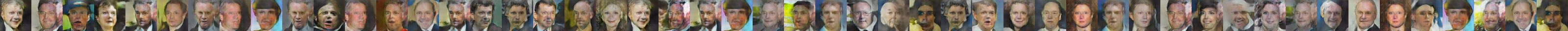}  & \underline{\bf 0.015} & 0.817 & 0.987\\
      \includegraphics[width=\textwidth,trim={1536px 0 0 0},clip]{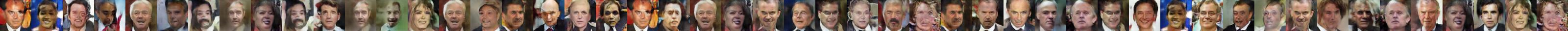}  & 0.020 & 0.784 & \underline{\bf 0.950}\\
      \includegraphics[width=\textwidth,trim={1536px 0 0 0},clip]{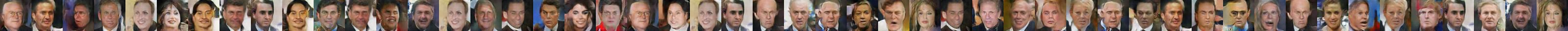}   & 0.024 & \underline{\bf 0.697} & 0.971\\\hline
  \end{tabular}}
  \caption{Results on GAN evaluation. Lower test statistics are best. Full results in Appendix~\ref{sec:app}.}
  \label{table:littlegans}
  \end{table}

  Since effective generative models will produce examples barely
  distinguishable from real data, two-sample tests arise as a natural
  alternative to evaluate generative models.
  Particularly, our interest is to evaluate the sample quality of
  generative models with intractable likelihoods, such as GANs
  \citep{goodfellow2014generative}.
  GANs implement the adversarial game
  \begin{equation}\label{eq:ganobj}
    \min_{g} \max_{d} \E_{x \sim P(X)}\, \left[\log(d(x))\right] + \E_{z \sim P(Z)} \left[\log(1-d(g(z)))\right],
  \end{equation}
  where $d(x)$ depicts the probability of the example $x$ 
  following the data distribution $P(X)$ versus being synthesized by the generator. This is
  according to a trainable \emph{discriminator} function $d$.
  In the adversarial game, the generator $g$ plays to fool the
  discriminator $d$ by transforming noise vectors $z \sim P(Z)$ into real-looking examples
  $g(z)$. On the opposite side, the discriminator
  plays to distinguish between real examples $x$ and synthesized examples $g(z)$.
  To approximate the solution to \eqref{eq:ganobj},
  alternate the optimization of the two losses \citep{goodfellow2014generative} given by
  \begin{align}
    L_d(d) &= \mathbb{E}_{x}\, \left[\ell(d(x), 1)\right] + \mathbb{E}_z\,  \left[ \ell(d(g(z)),0) \right],\nonumber\\
    L_g(g) &= \mathbb{E}_{x}\, \left[\ell(d(x), 0)\right] + \mathbb{E}_{z}\,\left[ \ell(d(g(z)),1) \right].\label{eq:ganobj2}
  \end{align}
  Under the formalization~\eqref{eq:ganobj2}, the adversarial game 
  reduces to the sequential minimization of $L_d(d)$ and $L_g(g)$, and reveals
  the true goal of the discriminator: to be the C2ST that
  best distinguishes data examples $x \sim P$ and synthesized examples $\hat{x}
  \sim \hat{P}$, where $\hat{P}$ is the probability distribution induced by
  sampling $z \sim P(Z)$ and computing $\hat{x} = g(z)$.  The
  formalization~\eqref{eq:ganobj2} unveils the existence of an arbitrary
  binary classification loss function $\ell$ (See Remark~\ref{remark:loss}),
  which in turn decides the divergence minimized between the real and fake
  data distributions \citep{nowozin2016f}.

  Unfortunately, the evaluation of the log-likelihood of a GANs is intractable. Therefore, we will employ a
  two-sample test to evaluate the quality of the fake examples $\hat{x} = g(z)$.
  In simple terms, evaluating a GAN in this
  manner amounts to withhold some real data from the training process, and use it later
  in a two-sample test against the same amount of synthesized
  data. When the two-sample test is a binary classifier (as discussed in
  Section~\ref{sec:neural}), this procedure is simply \emph{training a fresh
  discriminator on a fresh set of data}. Since we train and test this
  \emph{fresh} discriminator on held-out examples, it may differ from the
  discriminator trained along the GAN. In particular, the discriminator trained
  along with the GAN may have over-fitted to particular artifacts produced by
  the generator, thus becoming a poor C2ST.

  We evaluate the use of two-sample tests for model selection in GANs. 
  To this end, we train a number of DCGANs
  \citep{radford2015unsupervised} on the bedroom class of LSUN
  \citep{lsun} and the Labeled Faces in the Wild (LFW) dataset \citep{lfw}. We
  reused the Torch7 code of \citet{radford2015unsupervised} to train a set of
  DCGANs for $\{1,10,50,100,200\}$ epochs, where the generator and
  discriminator networks are convolutional neural networks
  \citep{lecun1998gradient} with $\{1,2,4,8\}\times\text{gf}$ and
  $\{1,2,4,8\}\times\text{df}$ filters per layer, respectively. We evaluate 
  each DCGAN on $10,000$ held-out examples using the fastest
  multi-dimensional two-sample tests: MMD, C2ST-NN, and C2ST-KNN.

  Our first experiments revealed an interesting result. When
  performing two-sample tests directly on pixels, all tests obtain
  near-perfect test accuracy when distinguishing between real and synthesized
  (fake) examples. Such near-perfect accuracy happens consistently across
  DCGANs, regardless of the visual quality of their examples. This is because,
  albeit visually appealing, the fake examples contain checkerboard-like
  artifacts that are sufficient for the tests to consistently
  differentiate between real and fake examples. 
  \citet{odena2016deconvolution} discovered this phenomenon concurrently with us. 
  
  On a second series of
  experiments, we featurize all images (both real and fake) using a deep
  convolutional ResNet \citep{he2015deep} pre-trained on ImageNet, a
  large dataset of natural images \citep{imagenet}. In particular, we use the
  \texttt{resnet-34} model from \citet{resnet}. Reusing a model pre-trained on
  natural images ensures that the test will distinguish between real and fake
  examples based only on natural image statistics, such as Gabor filters, edge
  detectors, and so on. Such a strategy is similar to perceptual
  losses \citep{perceptual} and inception scores \citep{salimans16}.
  In short, in order to evaluate how natural the images
  synthesized by a DCGAN look, one must employ a ``natural discriminator''.
  Table~\ref{table:littlegans} shows three GANs producing poor samples and
  three GANs producing good samples for the LSUN and LFW datasets, according to
  the MMD, C2ST-KNN, C2ST-NN tests on top of ResNet features. See
  Appendix~\ref{sec:app} for the full list of results. Although it is
  challenging to provide with an objective evaluation of our results, we
  believe that the rankings provided by two-sample tests could serve for efficient
  early stopping and model selection.

  \begin{remark}[How good is my GAN? Is it overfitting?]
  Evaluating generative models is a delicate issue \citep{theis2015note},
  but two-sample tests may offer some guidance.
  In particular, good (non-overfitting) generative models should produce similar two-sample test
  statistics when comparing their generated samples to both the train-set and the test-set samples.
  \footnote{As discussed with
  Arthur Gretton, if the generative model memorizes the train-set samples, a sufficiently
  large set of generated samples would reveal such memorization to the two-sample test. This
  is because some unique samples would appear multiple times in the set of generated samples,
  but not in the test-set of samples.}
  As a general recipe, prefer \emph{the smallest} (in number of parameters) generative model that achieves the \emph{same and small} two-sample
  test statistic when comparing their generated samples to both the train-set and test-set samples.
  
  We have seen that GANs of different quality may lead to the same
  (perfect) C2ST statistic.  To allow a finer comparison between generative
  models, we recommend implementing C2ST using a margin classifier with
  finite norm, or using as statistic the whole area under the C2ST
  training curve (on train-set or test-set samples).
  \end{remark}

  \subsection{Experiments on Interpretability}

  \begin{figure}
    \begin{center}
    \begin{subfigure}{0.26\textwidth}
      \includegraphics[width=\textwidth]{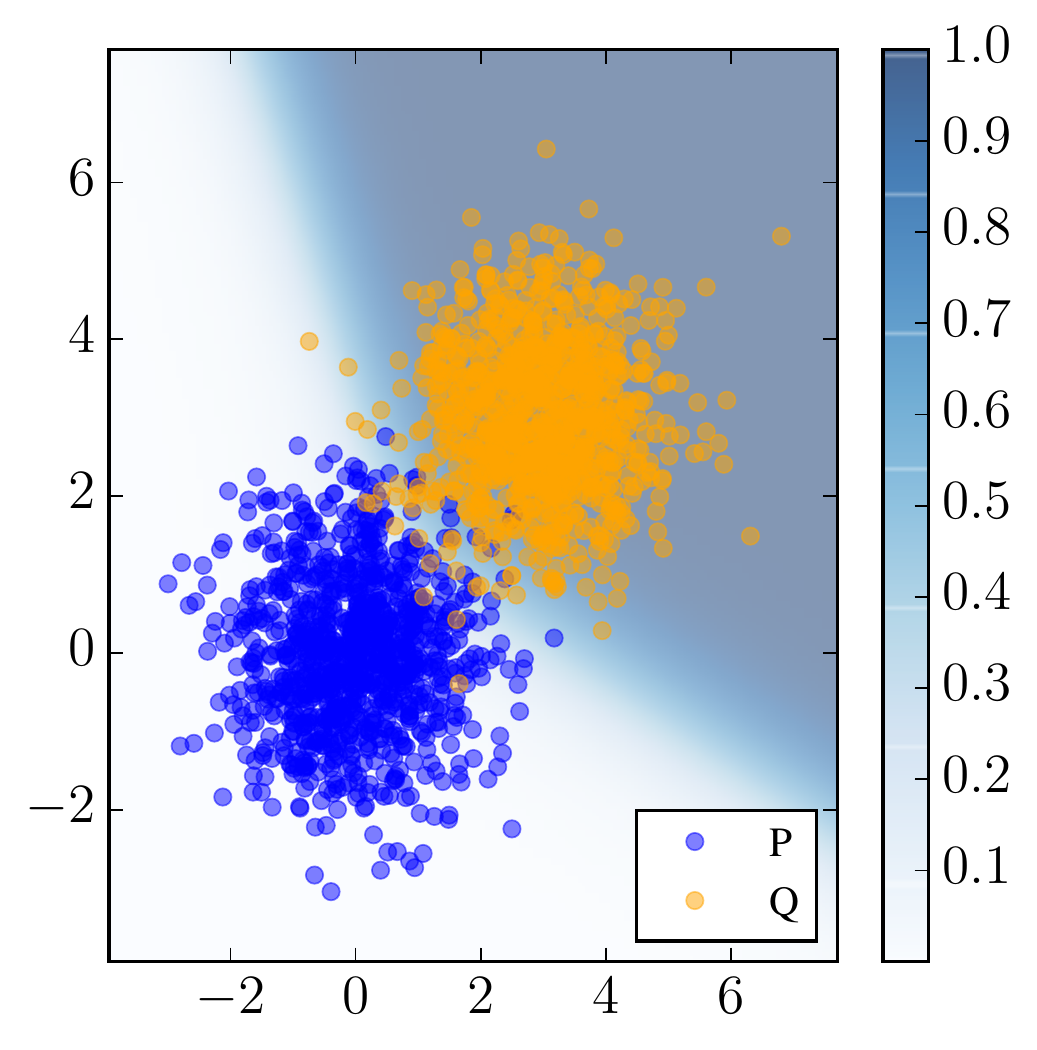}
    \end{subfigure}
    \hfill
    \begin{subfigure}{0.26\textwidth}
      \includegraphics[width=\textwidth]{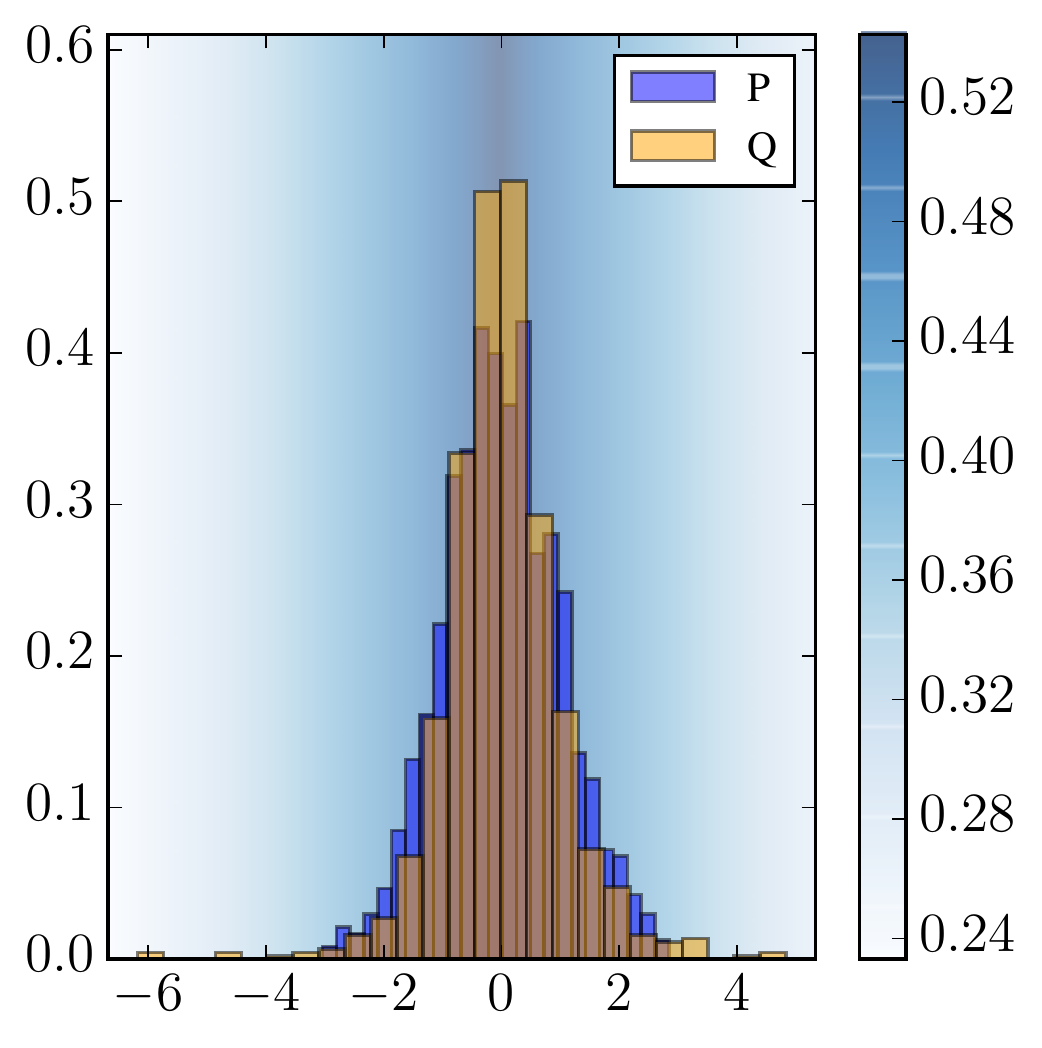}
    \end{subfigure}
    \hfill
    \begin{subfigure}{0.26\textwidth}
      \raisebox{0.07\height}{\includegraphics[width=\textwidth]{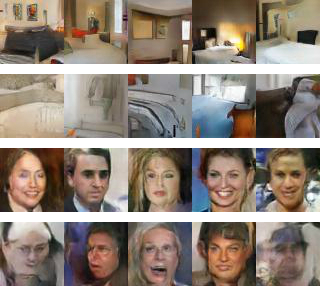}}
    \end{subfigure}
    \hfill
    \begin{subfigure}{0.18\textwidth}
      \includegraphics[width=0.32\textwidth]{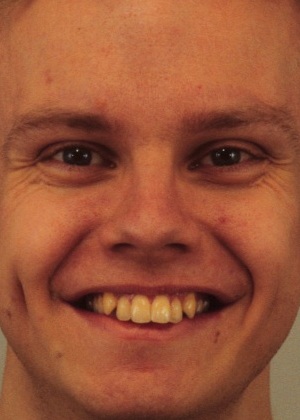}%
      \includegraphics[width=0.32\textwidth]{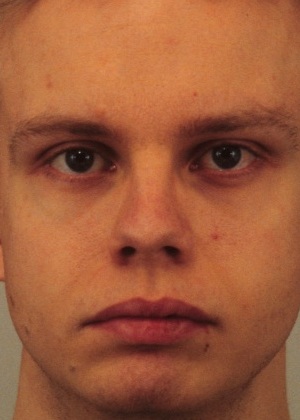}%
      \includegraphics[width=0.32\textwidth]{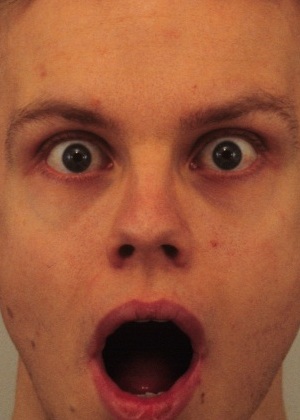}\\
      \includegraphics[width=0.32\textwidth]{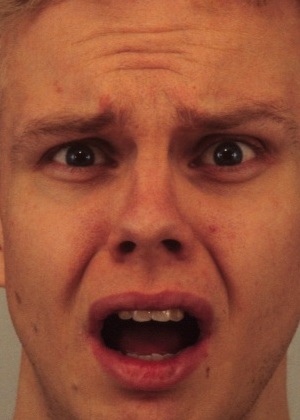}%
      \includegraphics[width=0.32\textwidth]{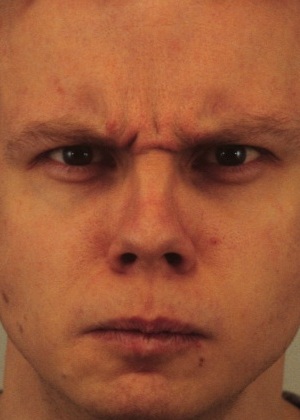}%
      \includegraphics[width=0.32\textwidth]{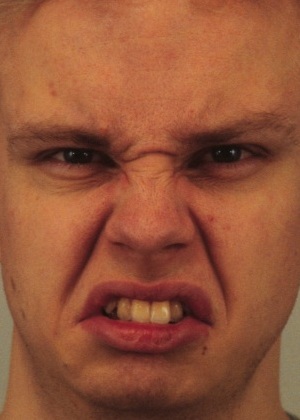}\\
      \includegraphics[width=0.32\textwidth]{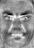}%
      \includegraphics[width=0.32\textwidth]{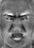}%
      \includegraphics[width=0.32\textwidth]{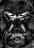}
    \end{subfigure}
    \end{center}
    \vspace{-0.2cm}
    \caption{Interpretability of C2ST. The color map corresponds to the value of $p(l=1|z)$.}
    \vspace{-0.2cm}
    \label{fig:interpretability}
  \end{figure}

  We illustrate the interpretability power of C2ST.
  First, the predictive uncertainty of C2ST sheds light on
  where the two samples under consideration agree or differ.
  In the context of GANs, this interpretability is useful to locate captured or dropped modes. 
  In the first plot
  of Figure~\ref{fig:interpretability}, a C2ST-NN separates two bivariate
  Gaussian
  distributions with different means. When performing this separation, the
  C2ST-NN provides an explicit decision boundary that illustrates \emph{where}
  the two distributions separate from each other. In the second plot of
  Figure~\ref{fig:interpretability}, a C2ST-NN separates a Gaussian
  distribution from a Student's t distribution with $\nu = 3$,
  after scaling both to zero-mean and unit-variance. The plot reveals that
  the peaks of the distributions are their most differentiating feature.
  Finally, the third plot of
  Figure~\ref{fig:interpretability} displays, for the LFW and LSUN datasets,
  five examples classified as real with high uncertainty (first row,
  better looking examples), and five examples classified as fake with
  high certainty (second row, worse looking examples).

  Second, the features learnt by the classifier of a C2ST 
  are also a mechanism to understand the differences between the two samples
  under study. The third plot of
  Figure~\ref{fig:interpretability} shows six examples from the Karolinska
  Directed Emotional Faces dataset, analyzed in Section~\ref{sec:smallexps}. In
  that same figure, we arrange the weights of the first linear layer of C2ST-NN
  into the feature most activated at positive examples
  (bottom left, positive facial expressions), the feature most activated at
  negative examples (bottom middle, negative facial expressions), and the
  ``discriminative feature'', obtained by substracting these two features
  (bottom right). The discriminative feature of C2ST-NN agrees with the one
  found by \citep{metests}: positive and negative facial expressions are best
  distinguished at the eyebrows, smile lines, and lips.  A similar analysis
  \cite{metests} on the C2ST-NN features in the NIPS article classification
  problem (Section~\ref{sec:smallexps}) reveals that the features most
  activated for the
  ``statistical learning theory'' category are those associated to the words
  \emph{inequ}, \emph{tight}, \emph{power}, \emph{sign}, \emph{hypothesi},
  \emph{norm}, \emph{hilbert}. The features most activated for
  the ``Bayesian inference'' category are those associated to the words
  \emph{infer}, \emph{markov}, \emph{graphic}, \emph{conjug}, \emph{carlo},
  \emph{automat}, \emph{laplac}.

  \section{Experiments on Conditional GANs for Causal Discovery}\label{sec:causality}

  In causal discovery, we study the causal structure underlying a set of $d$
  random variables $X_1, \ldots, X_d$. In particular, we assume that the random
  variables $X_1, \ldots, X_d$ share a causal structure described by a
  collection of Structural Equations, or SEs
  \citep{pearl2009causality}.  More specifically, we assume that the random
  variable $X_i$ takes values as described by the SE
  $X_i = g_i(\text{Pa}(X_i, \mathcal{G}), N_i)$,
  for all $i=1,\ldots,d$. In the previous, $\mathcal{G}$ is a Directed Acyclic
  Graph (DAG) with vertices associated to each of the random variables $X_1,
  \ldots, X_d$.  Also in the same equation, $\text{Pa}(X_i, \mathcal{G})$
  denotes the set of random variables which are parents of $X_i$ in the graph
  $\mathcal{G}$, and $N_i$ is an independent noise random variable that
  follows the probability distribution $P(N_i)$. Then, we say that $X_i \to X_j$
  if $X_i \in \text{Pa}(X_j)$, since a change in $X_i$ will \emph{cause} a change in $X_j$,
  as described by the $i$-th SE.

  The goal of causal discovery is to infer the causal graph $\mathcal{G}$ given
  a sample from $P(X_1, \ldots, X_d)$. For
  the sake of simplicity, we focus on the discovery of causal relations
  between two random variables, denoted by $X$ and $Y$. That is, given the sample
  $\mathcal{D} = \{(x_i,y_i)\}_{i=1}^n \sim P^n(X,Y)$, our goal is to
  conclude whether ``$X$ causes $Y$'', or ``$Y$
  causes $X$''. We call this problem \emph{cause-effect discovery}
  \citep{mooij2014distinguishing}. In the case where $X \to Y$, we can write
  the cause-effect relationship as:
  \begin{align}
    x \sim P(X),\quad
    n \sim P(N),\quad
    y \leftarrow g(x,n).\label{eq:mechanism}
  \end{align}
  The current state-of-the-art in the cause-effect discovery is the family of
  Additive Noise Models, or ANM \citep{mooij2014distinguishing}. These methods
  assume that the SE \eqref{eq:mechanism} allow the
  expression $y \leftarrow g(x)+n$, and exploit the independence assumption
  between the cause random variable $X$ and the noise random variable $N$ to
  analyze the distribution of nonlinear regression residuals, in both
  causal directions.

  Unfortunately, assuming independent additive noise is often too simplistic
  (for instance, the noise could be heteroskedastic or multiplicative).
  Because of this reason, we propose to use Conditional Generative Adversarial
  Networks, or CGANs \citep{conditionalGans} to address the problem of
  cause-effect discovery. Our motivation is the shocking resemblance between
  the generator
  of a CGAN and the
  SE \eqref{eq:mechanism}: the random variable $X$ is the
  conditioning variable
  input to the generator, the random variable $N$ is the noise variable input
  to the generator, and the random variable $Y$ is the variable synthesized by
  the generator. Furthermore, CGANs respect the independence between the cause
  $X$ and the noise $N$ by construction, since $n \sim P(N)$ is independent
  from all other variables.  This way, CGANs bypass the additive noise
  assumption naturally, and allow arbitrary interactions $g(X,N)$
  between the cause variable $X$ and the noise variable $N$.

  To implement our cause-effect inference algorithm in practice,
  recall that training a CGAN from $X$ to $Y$ minimizes the two
  following objectives in alternation:
  \begin{align}
    L_d(d) &= \mathbb{E}_{x,y}\, \left[\ell(d(x,y), 1)\right] + \mathbb{E}_{x,z}\, \left[\ell(d(x,g(x,z)), 0)\right],\nonumber\\
    L_g(g) &= \mathbb{E}_{x,y}\, \left[\ell(d(x,y), 0)\right] + \mathbb{E}_{x,z}\, \left[\ell(d(x,g(x,z)),1)\right].\nonumber
  \end{align}
  Our recipe for cause-effect is to learn two CGANs: one with a
  generator $g_y$ from $X$ to $Y$ to synthesize the dataset $\mathcal{D}_{X\to
  Y} = \{(x_i, g_y(x_i,z_i))\}_{i=1}^n$, and one with a generator $g_x$ from $Y$
  to $X$ to synthesize the dataset $\mathcal{D}_{X\leftarrow Y} =
  \{(g_x(y_i,z_i),y_i)\}_{i=1}^n$. Then, we prefer the causal direction
  $X \to Y$ if the two-sample test statistic between the real sample
  $\mathcal{D}$ and $\mathcal{D}_{X\to Y}$ is smaller than the one
  between $\mathcal{D}$ and $\mathcal{D}_{Y \to X}$. Thus, our method is
  Occam's razor at play: declare the simplest direction (in terms of
  conditional generative modeling) as the true causal direction.

  \begin{table}
    \resizebox{\textwidth}{!}{
    \begin{tabular}{|l|c|c|c|c|c|c|}
    \hline
    Method & ANM-HSIC & IGCI & RCC & CGAN-C2ST & Ensemble & C2ST type\\
    \hline
    \multirow{3}{*}{Accuracy} & \multirow{3}{*}{$67\percent$} & \multirow{3}{*}{$71\percent$} & \multirow{3}{*}{$76\percent$} & $73\percent$ & \textbf{82\percent} & KNN\\\cline{5-6}
             & & & & $70\percent$ & $73\percent$ & NN\\\cline{5-6}
             & & & & $58\percent$ & $65\percent$ & MMD\\\hline
    \end{tabular}
    }
    \vspace{-0.2cm}
    \caption{Results on cause-effect discovery on the T\"ubingen pairs experiment.}
    \vspace{-0.2cm}
    \label{table:tuebingen}
  \end{table}

  Table~\ref{table:tuebingen} summarizes the performance of this procedure when
  applied to the $99$ T\"ubingen cause-effect pairs dataset, version August
  2016 \citep{mooij2014distinguishing}. RCC is the Randomized Causation
  Coefficient of \citep{dlp-clt}. The Ensemble-CGAN-C2ST trains 100 CGANs, and
  decides the causal direction by comparing the top generator obtained in each
  causal direction, as told by C2ST-KNN. The need to ensemble is a remainder of
  the unstable behaviour of generative adversarial training, but also
  highlights the promise of such models for causal discovery.

  \section{Conclusion}
  Our \emph{take-home message} is that modern binary classifiers
  can be easily turned into powerful two-sample tests. We have shown that these
  \emph{classifier two-sample tests} set a new state-of-the-art in performance,
  and enjoy unique attractive properties: they are easy to implement, learn a
  representation of the data on the fly, have simple asymptotic distributions,
  and allow different ways to interpret how the two samples under study differ.
  Looking into the future, the use of binary classifiers as two-sample tests
  provides a flexible and scalable approach for the evaluation and comparison
  of generative models (such as GANs), and opens the door to novel applications of these
  methods, such as causal discovery.
  
  \bibliography{classifier_tests}
  \bibliographystyle{iclr2017_conference}

  \clearpage
  \newpage
  \appendix
  \section{Results on Evaluation of Generative Adversarial Networks}\label{sec:app}
  \begin{table}[h!]
    \begin{center}
    \resizebox{0.99\textwidth}{!}{
    \begin{tabular}{lllcccc}
      gf & df & ep & random sample & MMD & KNN & NN \\
      \hline
      -  & - & - & \includegraphics[width=\textwidth,trim={1536px 0 0 0},clip]{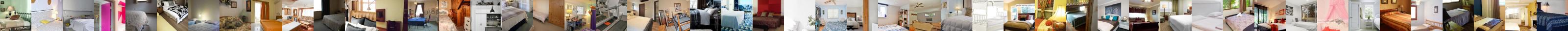} & - & - & -\\
      \hline
      32 & 32 & 1   & \includegraphics[width=\textwidth,trim={1536px 0 0 0},clip]{figures/bedrooms_g32_d32_ep1_generator.jpg}    & 0.154 & 0.994 & 1.000\\
      32 & 32 & 10  & \includegraphics[width=\textwidth,trim={1536px 0 0 0},clip]{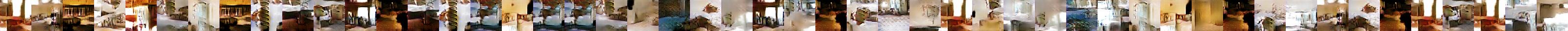}   & 0.024 & 0.831 & 0.996\\
      32 & 32 & 50  & \includegraphics[width=\textwidth,trim={1536px 0 0 0},clip]{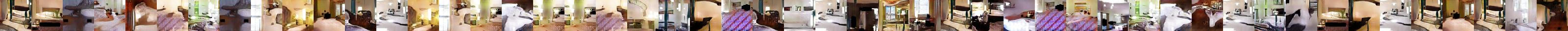}   & 0.026 & 0.758 & 0.983\\
      32 & 32 & 100 & \includegraphics[width=\textwidth,trim={1536px 0 0 0},clip]{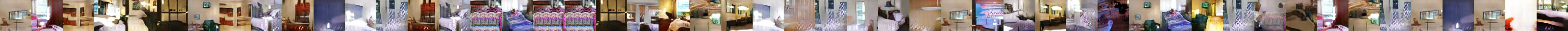}  & 0.014 & 0.797 & 0.974\\
      32 & 32 & 200 & \includegraphics[width=\textwidth,trim={1536px 0 0 0},clip]{figures/bedrooms_g32_d32_ep200_generator.jpg}  & \underline{\bf 0.012} & 0.798 & 0.964\\
      \hline
      32 & 64 & 1   & \includegraphics[width=\textwidth,trim={1536px 0 0 0},clip]{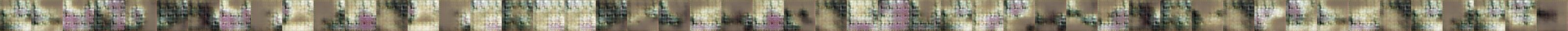}    & 0.330 & 0.984 & 1.000\\
      32 & 64 & 10  & \includegraphics[width=\textwidth,trim={1536px 0 0 0},clip]{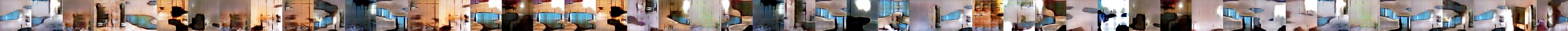}   & 0.035 & 0.897 & 0.997\\
      32 & 64 & 50  & \includegraphics[width=\textwidth,trim={1536px 0 0 0},clip]{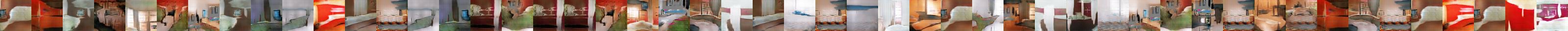}   & 0.020 & 0.804 & 0.989\\
      32 & 64 & 100 & \includegraphics[width=\textwidth,trim={1536px 0 0 0},clip]{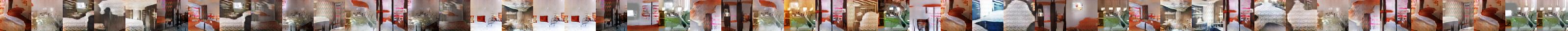}  & 0.032 & 0.936 & 0.998\\
      32 & 64 & 200 & \includegraphics[width=\textwidth,trim={1536px 0 0 0},clip]{figures/bedrooms_g32_d64_ep200_generator.jpg}  & 0.048 & 0.962 & 1.000\\
      \hline
      32 & 96 & 1   & \includegraphics[width=\textwidth,trim={1536px 0 0 0},clip]{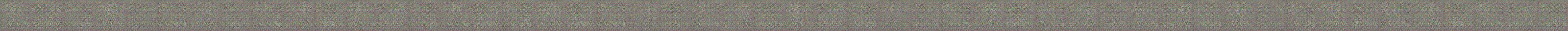}    & 0.915 & 0.997 & 1.000\\
      32 & 96 & 10  & \includegraphics[width=\textwidth,trim={1536px 0 0 0},clip]{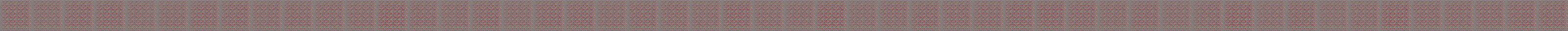}   & 0.927 & 0.991 & 1.000\\
      32 & 96 & 50  & \includegraphics[width=\textwidth,trim={1536px 0 0 0},clip]{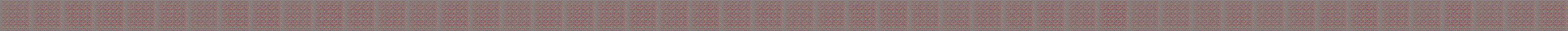}   & 0.924 & 0.991 & 1.000\\
      32 & 96 & 100 & \includegraphics[width=\textwidth,trim={1536px 0 0 0},clip]{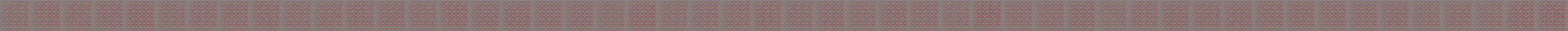}  & 0.928 & 0.991 & 1.000\\
      32 & 96 & 200 & \includegraphics[width=\textwidth,trim={1536px 0 0 0},clip]{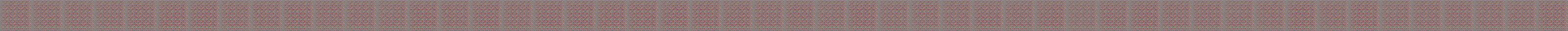}  & 0.928 & 0.991 & 1.000\\
      \hline
      64 & 32 & 1   & \includegraphics[width=\textwidth,trim={1536px 0 0 0},clip]{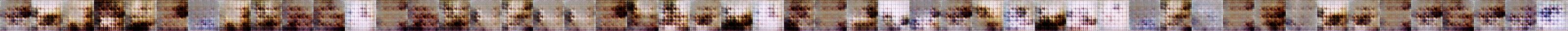}    & 0.389 & 0.987 & 1.000\\
      64 & 32 & 10  & \includegraphics[width=\textwidth,trim={1536px 0 0 0},clip]{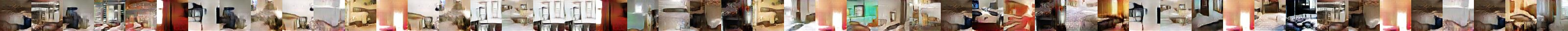}   & 0.023 & 0.842 & 0.979\\
      64 & 32 & 50  & \includegraphics[width=\textwidth,trim={1536px 0 0 0},clip]{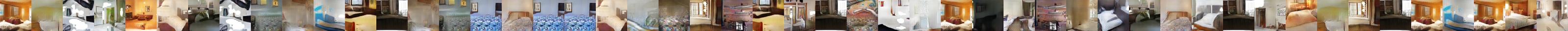}   & 0.018 & 0.788 & 0.977\\
      64 & 32 & 100 & \includegraphics[width=\textwidth,trim={1536px 0 0 0},clip]{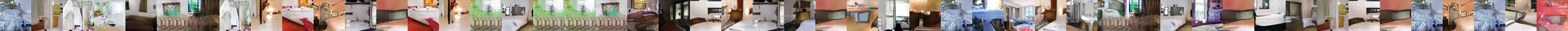}  & 0.017 & 0.753 & 0.959\\
      64 & 32 & 200 & \includegraphics[width=\textwidth,trim={1536px 0 0 0},clip]{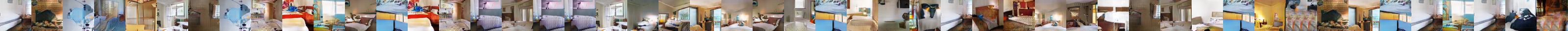}  & 0.018 & 0.736 & 0.963\\
      \hline
      64 & 64 & 1   & \includegraphics[width=\textwidth,trim={1536px 0 0 0},clip]{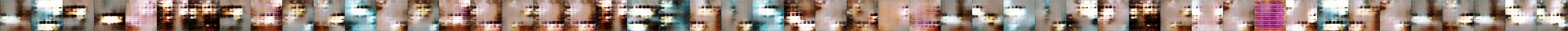}    & 0.313 & 0.964 & 1.000\\
      64 & 64 & 10  & \includegraphics[width=\textwidth,trim={1536px 0 0 0},clip]{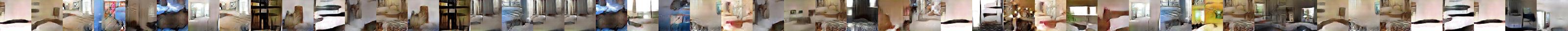}   & 0.021 & 0.825 & 0.988\\
      64 & 64 & 50  & \includegraphics[width=\textwidth,trim={1536px 0 0 0},clip]{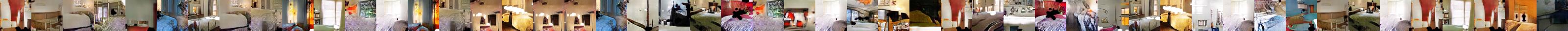}   & 0.014 & 0.864 & 0.978\\
      64 & 64 & 100 & \includegraphics[width=\textwidth,trim={1536px 0 0 0},clip]{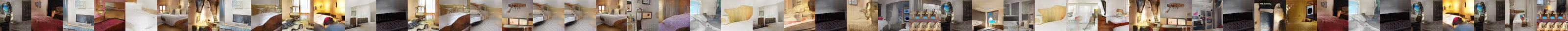}  & 0.019 & 0.685 & 0.978\\
      64 & 64 & 200 & \includegraphics[width=\textwidth,trim={1536px 0 0 0},clip]{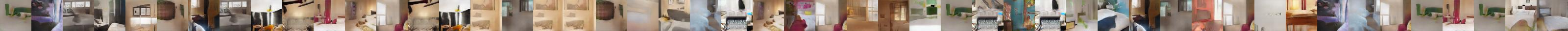}  & 0.021 & 0.775 & 0.980\\
      \hline
      64 & 96 & 1   & \includegraphics[width=\textwidth,trim={1536px 0 0 0},clip]{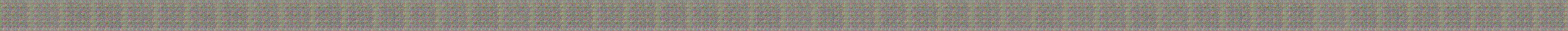}    & 0.891 & 0.996 & 1.000\\
      64 & 96 & 10  & \includegraphics[width=\textwidth,trim={1536px 0 0 0},clip]{figures/bedrooms_g64_d96_ep10_generator.jpg}   & 0.158 & 0.830 & 0.999\\
      64 & 96 & 50  & \includegraphics[width=\textwidth,trim={1536px 0 0 0},clip]{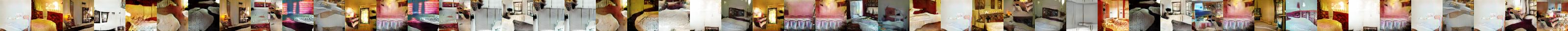}   & 0.015 & 0.801 & 0.980\\
      64 & 96 & 100 & \includegraphics[width=\textwidth,trim={1536px 0 0 0},clip]{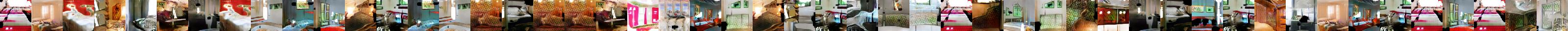}  & 0.016 & 0.866 & 0.976\\
      64 & 96 & 200 & \includegraphics[width=\textwidth,trim={1536px 0 0 0},clip]{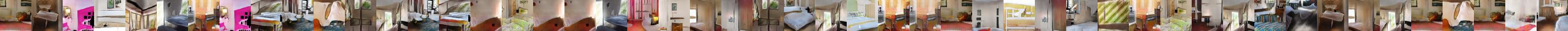}  & 0.020 & 0.755 & 0.983\\
      \hline
      96 & 32 & 1   & \includegraphics[width=\textwidth,trim={1536px 0 0 0},clip]{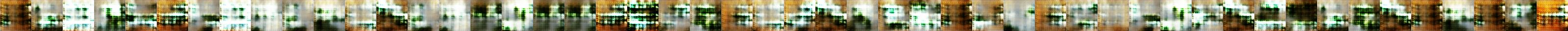}    & 0.356 & 0.986 & 1.000\\
      96 & 32 & 10  & \includegraphics[width=\textwidth,trim={1536px 0 0 0},clip]{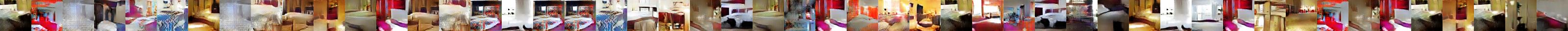}   & 0.022 & 0.770 & 0.991\\
      96 & 32 & 50  & \includegraphics[width=\textwidth,trim={1536px 0 0 0},clip]{figures/bedrooms_g96_d32_ep50_generator.jpg}   & 0.024 & 0.748 & \underline{\bf 0.949}\\
      96 & 32 & 100 & \includegraphics[width=\textwidth,trim={1536px 0 0 0},clip]{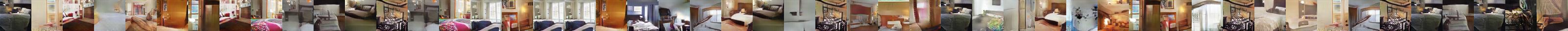}  & 0.022 & 0.745 & 0.965\\
      96 & 32 & 200 & \includegraphics[width=\textwidth,trim={1536px 0 0 0},clip]{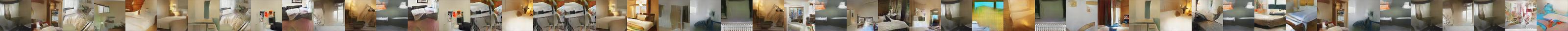}  & 0.024 & 0.689 & 0.981\\
      \hline
      96 & 64 & 1   & \includegraphics[width=\textwidth,trim={1536px 0 0 0},clip]{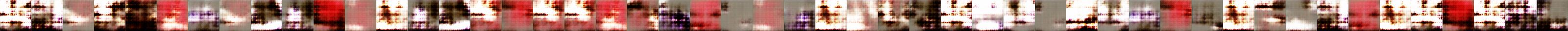}    & 0.287 & 0.978 & 1.000\\
      96 & 64 & 10  & \includegraphics[width=\textwidth,trim={1536px 0 0 0},clip]{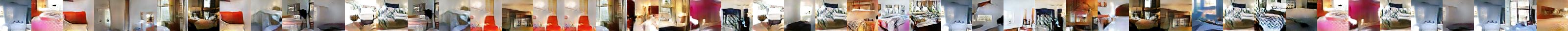}   & \underline{\bf 0.012} & 0.825 & \underline{\bf 0.966}\\
      96 & 64 & 50  & \includegraphics[width=\textwidth,trim={1536px 0 0 0},clip]{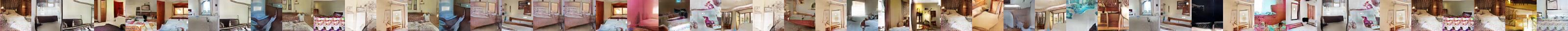}   & 0.017 & 0.812 & 0.962\\
      96 & 64 & 100 & \includegraphics[width=\textwidth,trim={1536px 0 0 0},clip]{figures/bedrooms_g96_d64_ep100_generator.jpg}  & 0.019 & \underline{\bf 0.670} & 0.983\\
      96 & 64 & 200 & \includegraphics[width=\textwidth,trim={1536px 0 0 0},clip]{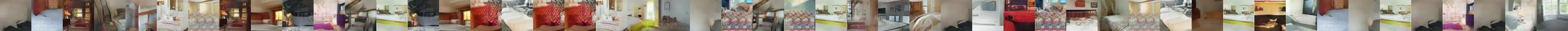}  & 0.020 & 0.711 & 0.972\\
      \hline
      96 & 96 & 1   & \includegraphics[width=\textwidth,trim={1536px 0 0 0},clip]{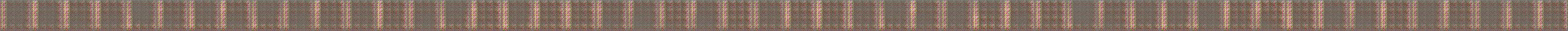}    & 0.672 & 0.999 & 1.000\\
      96 & 96 & 10  & \includegraphics[width=\textwidth,trim={1536px 0 0 0},clip]{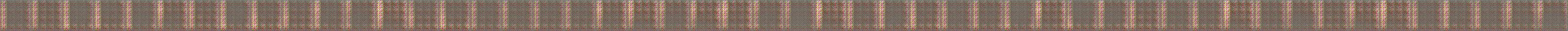}   & 0.671 & 0.999 & 1.000\\
      96 & 96 & 50  & \includegraphics[width=\textwidth,trim={1536px 0 0 0},clip]{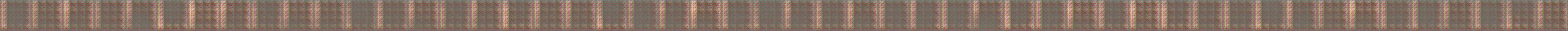}   & 0.829 & 0.999 & 1.000\\
      96 & 96 & 100 & \includegraphics[width=\textwidth,trim={1536px 0 0 0},clip]{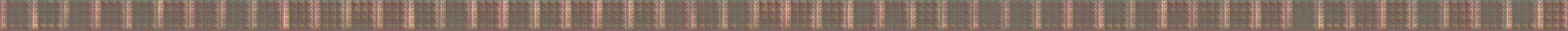}  & 0.668 & 0.999 & 1.000\\
      96 & 96 & 200 & \includegraphics[width=\textwidth,trim={1536px 0 0 0},clip]{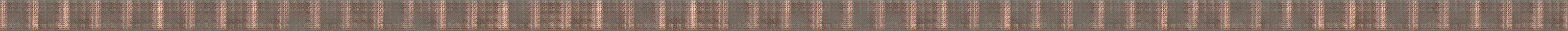}  & 0.849 & 0.999 & 1.000\\
      \hline
    \end{tabular}
    }
    \end{center}
    \vspace{-0.2cm}
    \caption{GAN evaluation results on the LSUN dataset, for all epochs (ep),
    filters in discriminator (df), filters in generator (gf), and test
    statistics (for MMD, C2ST-KNN, C2ST-NN). A lower test statistic estimates
    that the GAN produces better samples. Best viewed with zoom.}
    \vspace{-0.2cm}
    \label{table:bedrooms}
  \end{table}
  \clearpage
  \newpage

  \begin{table}[h!]
    \begin{center}
    \resizebox{0.99\textwidth}{!}{
    \begin{tabular}{lllcccc}
      gf & df & ep & random sample & MMD & KNN & NN \\
      \hline
      -  & - & - & \includegraphics[width=\textwidth,trim={1536px 0 0 0},clip]{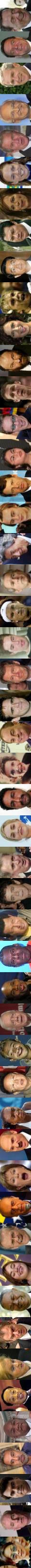} & - & - & -\\
      \hline
      32 & 32 & 1   & \includegraphics[width=\textwidth,trim={1536px 0 0 0},clip]{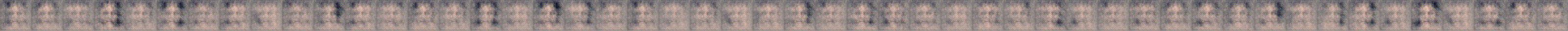}    & 0.806 & 1.000 & 1.000\\
      32 & 32 & 10  & \includegraphics[width=\textwidth,trim={1536px 0 0 0},clip]{figures/faces_g32_d32_ep10_generator.jpg}   & 0.152 & 0.940 & 1.000\\
      32 & 32 & 50  & \includegraphics[width=\textwidth,trim={1536px 0 0 0},clip]{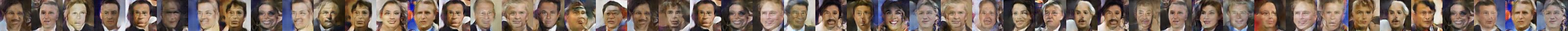}   & 0.042 & 0.788 & 0.993\\
      32 & 32 & 100 & \includegraphics[width=\textwidth,trim={1536px 0 0 0},clip]{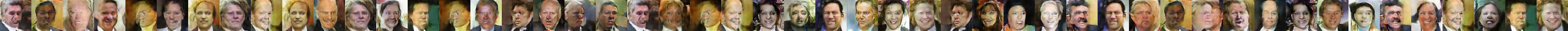}  & 0.029 & 0.808 & 0.982\\
      32 & 32 & 200 & \includegraphics[width=\textwidth,trim={1536px 0 0 0},clip]{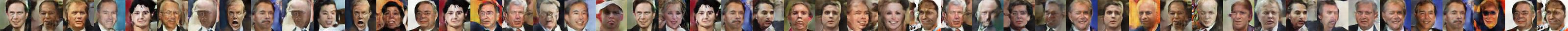}  & 0.022 & 0.776 & 0.970\\
      \hline
      32 & 64 & 1   & \includegraphics[width=\textwidth,trim={1536px 0 0 0},clip]{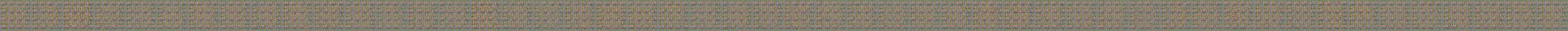}    & 0.994 & 1.000 & 1.000\\
      32 & 64 & 10  & \includegraphics[width=\textwidth,trim={1536px 0 0 0},clip]{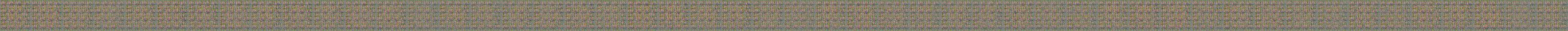}   & 0.989 & 1.000 & 1.000\\
      32 & 64 & 50  & \includegraphics[width=\textwidth,trim={1536px 0 0 0},clip]{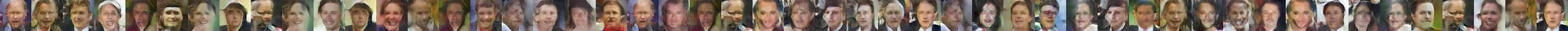}   & 0.050 & 0.808 & 0.985\\
      32 & 64 & 100 & \includegraphics[width=\textwidth,trim={1536px 0 0 0},clip]{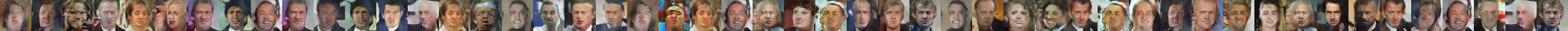}  & 0.036 & 0.766 & 0.972\\
      32 & 64 & 200 & \includegraphics[width=\textwidth,trim={1536px 0 0 0},clip]{figures/faces_g32_d64_ep200_generator.jpg}  & \underline{\bf 0.015} & 0.817 & 0.987\\
      \hline
      32 & 96 & 1   & \includegraphics[width=\textwidth,trim={1536px 0 0 0},clip]{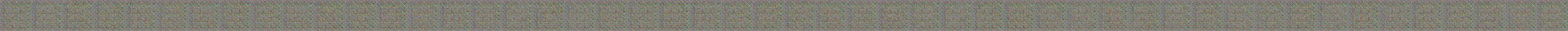}    & 0.995 & 1.000 & 1.000\\
      32 & 96 & 10  & \includegraphics[width=\textwidth,trim={1536px 0 0 0},clip]{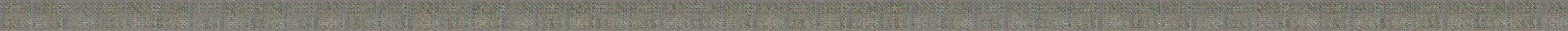}   & 0.992 & 1.000 & 1.000\\
      32 & 96 & 50  & \includegraphics[width=\textwidth,trim={1536px 0 0 0},clip]{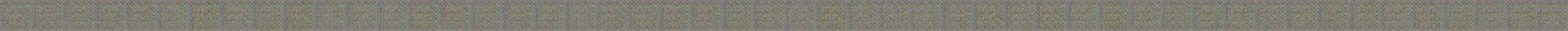}   & 0.995 & 1.000 & 1.000\\
      32 & 96 & 100 & \includegraphics[width=\textwidth,trim={1536px 0 0 0},clip]{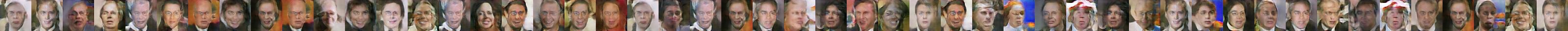}  & 0.053 & 0.778 & 0.987\\
      64 & 96 & 200 & \includegraphics[width=\textwidth,trim={1536px 0 0 0},clip]{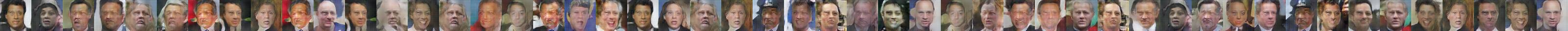}  & 0.037 & 0.779 & 0.995\\
      \hline
      64 & 32 & 1   & \includegraphics[width=\textwidth,trim={1536px 0 0 0},clip]{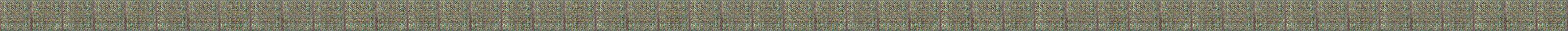}    & 1.041 & 1.000 & 1.000\\
      64 & 32 & 10  & \includegraphics[width=\textwidth,trim={1536px 0 0 0},clip]{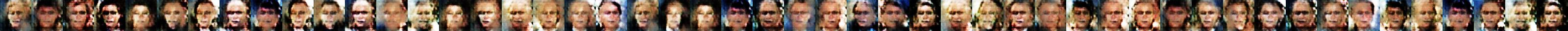}   & 0.086 & 0.971 & 1.000\\
      64 & 32 & 50  & \includegraphics[width=\textwidth,trim={1536px 0 0 0},clip]{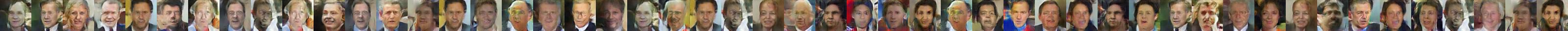}   & 0.043 & 0.756 & 0.988\\
      64 & 32 & 100 & \includegraphics[width=\textwidth,trim={1536px 0 0 0},clip]{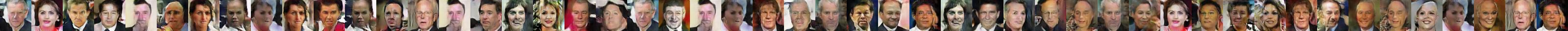}  & 0.018 & 0.746 & 0.973\\
      64 & 32 & 200 & \includegraphics[width=\textwidth,trim={1536px 0 0 0},clip]{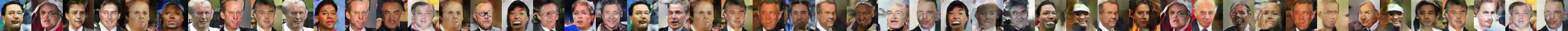}  & 0.025 & 0.757 & 0.972\\
      \hline
      64 & 64 & 1   & \includegraphics[width=\textwidth,trim={1536px 0 0 0},clip]{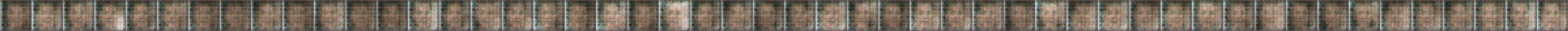}    & 0.836 & 1.000 & 1.000\\
      64 & 64 & 10  & \includegraphics[width=\textwidth,trim={1536px 0 0 0},clip]{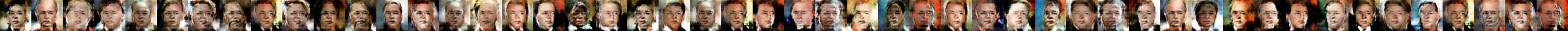}   & 0.103 & 0.910 & 0.998\\
      64 & 64 & 50  & \includegraphics[width=\textwidth,trim={1536px 0 0 0},clip]{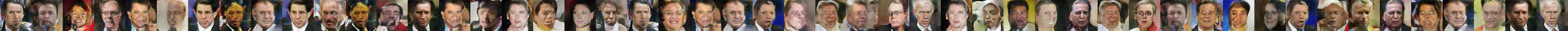}   & 0.018 & 0.712 & 0.973\\
      64 & 64 & 100 & \includegraphics[width=\textwidth,trim={1536px 0 0 0},clip]{figures/faces_g64_d64_ep100_generator.jpg}  & 0.020 & 0.784 & \underline{\bf 0.950}\\
      64 & 64 & 200 & \includegraphics[width=\textwidth,trim={1536px 0 0 0},clip]{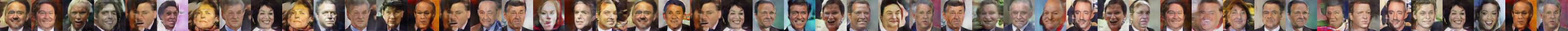}  & 0.022 & 0.719 & 0.974\\
      \hline
      64 & 96 & 1   & \includegraphics[width=\textwidth,trim={1536px 0 0 0},clip]{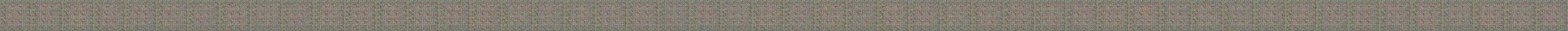}    & 1.003 & 1.000 & 1.000\\
      64 & 96 & 10  & \includegraphics[width=\textwidth,trim={1536px 0 0 0},clip]{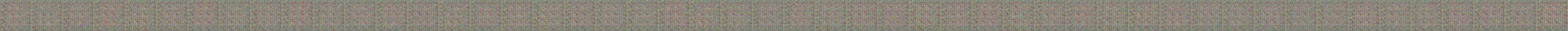}   & 1.015 & 1.000 & 1.000\\
      64 & 96 & 50  & \includegraphics[width=\textwidth,trim={1536px 0 0 0},clip]{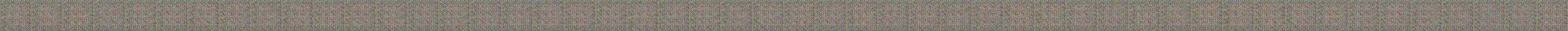}   & 1.002 & 1.000 & 1.000\\
      64 & 96 & 100 & \includegraphics[width=\textwidth,trim={1536px 0 0 0},clip]{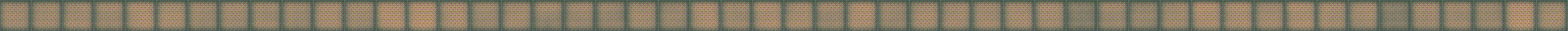}  & 1.063 & 1.000 & 1.000\\
      64 & 96 & 200 & \includegraphics[width=\textwidth,trim={1536px 0 0 0},clip]{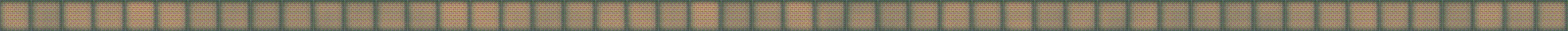}  & 1.061 & 1.000 & 1.000\\
      \hline
      96 & 32 & 1   & \includegraphics[width=\textwidth,trim={1536px 0 0 0},clip]{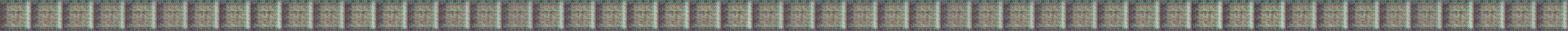}    & 1.022 & 1.000 & 1.000\\
      96 & 32 & 10  & \includegraphics[width=\textwidth,trim={1536px 0 0 0},clip]{figures/faces_g96_d32_ep10_generator.jpg}   & 0.222 & 0.978 & 1.000\\
      96 & 32 & 50  & \includegraphics[width=\textwidth,trim={1536px 0 0 0},clip]{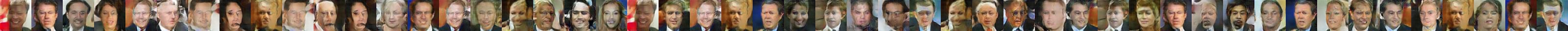}   & 0.026 & 0.734 & 0.965\\
      96 & 32 & 100 & \includegraphics[width=\textwidth,trim={1536px 0 0 0},clip]{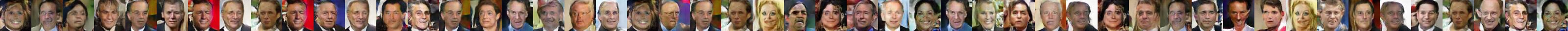}  & 0.016 & 0.735 & 0.964\\
      96 & 32 & 200 & \includegraphics[width=\textwidth,trim={1536px 0 0 0},clip]{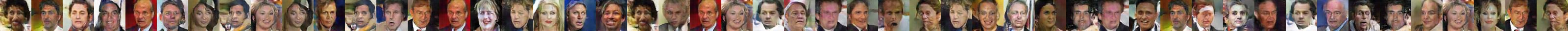}  & 0.021 & 0.780 & 0.973\\
      \hline
      96 & 64 & 1   & \includegraphics[width=\textwidth,trim={1536px 0 0 0},clip]{figures/faces_g96_d64_ep1_generator.jpg}    & 0.715 & 1.000 & 1.000\\
      96 & 64 & 10  & \includegraphics[width=\textwidth,trim={1536px 0 0 0},clip]{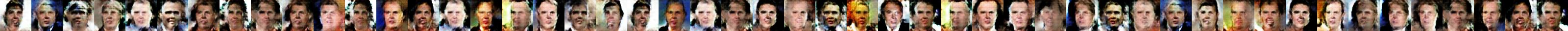}   & 0.042 & 0.904 & 0.999\\
      96 & 64 & 50  & \includegraphics[width=\textwidth,trim={1536px 0 0 0},clip]{figures/faces_g96_d64_ep50_generator.jpg}   & 0.024 & \underline{\bf 0.697} & 0.971\\
      96 & 64 & 100 & \includegraphics[width=\textwidth,trim={1536px 0 0 0},clip]{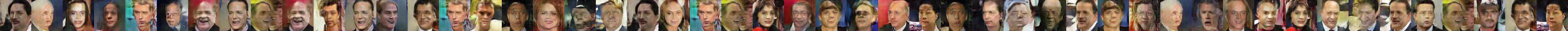}  & 0.028 & 0.744 & 0.983\\
      96 & 64 & 200 & \includegraphics[width=\textwidth,trim={1536px 0 0 0},clip]{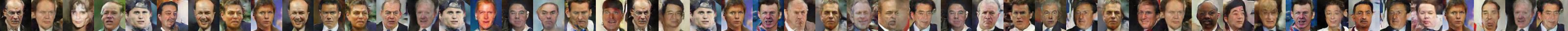}  & 0.020 & 0.697 & 0.976\\
      \hline
      96 & 96 & 1   & \includegraphics[width=\textwidth,trim={1536px 0 0 0},clip]{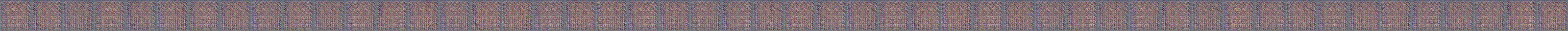}    & 0.969 & 1.000 & 1.000\\
      96 & 96 & 10  & \includegraphics[width=\textwidth,trim={1536px 0 0 0},clip]{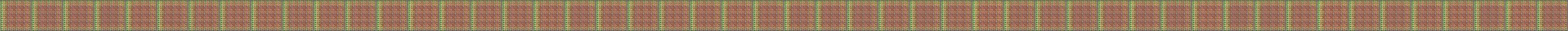}   & 0.920 & 1.000 & 1.000\\
      96 & 96 & 50  & \includegraphics[width=\textwidth,trim={1536px 0 0 0},clip]{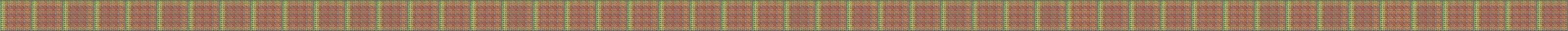}   & 0.926 & 1.000 & 1.000\\
      96 & 96 & 100 & \includegraphics[width=\textwidth,trim={1536px 0 0 0},clip]{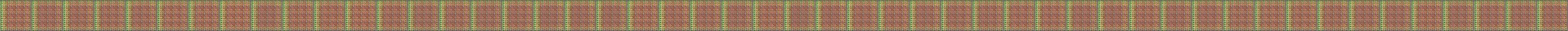}  & 0.920 & 1.000 & 1.000\\
      96 & 96 & 200 & \includegraphics[width=\textwidth,trim={1536px 0 0 0},clip]{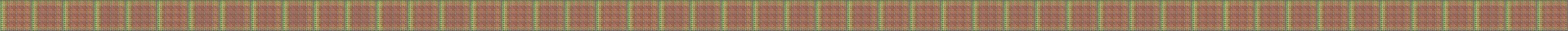}  & 0.923 & 1.000 & 1.000\\
      \hline
    \end{tabular}
    }
    \end{center}
    \vspace{-0.2cm}
    \caption{GAN evaluation results on the LFW dataset, for all epochs (ep),
    filters in discriminator (df), filters in generator (gf), and test
    statistics (for MMD, C2ST-KNN, C2ST-NN). A lower test statistic estimates
    that the GAN produces better samples. Best viewed with zoom.}
    \vspace{-0.2cm}
    \label{table:faces}
  \end{table}
  \clearpage
  \newpage
 
  \section{Proof of Theorem~\ref{thm:thm1}}\label{app:proof}
  Our statistic is a random variable $T \sim
  \mathcal{N}\left(\frac{1}{2}, \frac{1}{4 n_\text{te}}\right)$ under the null
  hypothesis, and $T \sim \mathcal{N}\left(\frac{1}{2} + \epsilon,
  n_\text{te}^{-1}\left(\frac{1}{4}-\epsilon^2\right)\right)$ under the
  alternative hypothesis.  Furthermore, at a significance level $\alpha$, the
  threshold of our statistic is $z_\alpha = \frac{1}{2} +
  \frac{\Phi^{-1}(1-\alpha)}{\sqrt{4n_\text{te}}}$; under this threshold we
  would accept the null hypothesis. Then, the probability of making a type-II error is 
  \begin{align*}
    \mathbb{P}_{T\sim \mathcal{N}\left(\frac{1}{2}+\epsilon,
    \frac{\frac{1}{4}-\epsilon^2}{n_\text{te}}\right)}\left(T < z_\alpha \right) &=
    \mathbb{P}_{T'\sim \mathcal{N}\left(0,
    \frac{\frac{1}{4}-\epsilon^2}{n_\text{te}}\right)}\left(T' <
    \frac{\Phi^{-1}(1-\alpha)}{\sqrt{4n_\text{te}}}-\epsilon\right)\\
    &=\Phi\left(\sqrt{\frac{n_\text{te}}{\frac{1}{4}-\epsilon^2}}
    \left(
    \frac{\Phi^{-1}(1-\alpha)}{\sqrt{4n_\text{te}}}-\epsilon\right)\right)\\
    &=\Phi
    \left(
    \frac{\Phi^{-1}(1-\alpha)/2 - \epsilon\sqrt{n_\text{te}}}{\sqrt{\frac{1}{4}-\epsilon^2}}
    \right).
  \end{align*}
  Therefore, the power of the test is
  \begin{equation*}
    \pi(\alpha, n_\text{te}, \epsilon) = 1-\Phi
    \left(
    \frac{\Phi^{-1}(1-\alpha)/2 - \epsilon\sqrt{n_\text{te}}}{\sqrt{\frac{1}{4}-\epsilon^2}}
    \right) =
    \Phi
    \left(
    \frac{\epsilon\sqrt{n_\text{te}}-\Phi^{-1}(1-\alpha)/2}{\sqrt{\frac{1}{4}-\epsilon^2}}
    \right),
  \end{equation*}
  which concludes the proof.
  
  \section{Acknowledgements}
 
   We are thankful to L. Bottou, B. Graham, D. Kiela, M.  Rojas-Carulla, I.
   Tolstikhin, and M. Tygert for their help in improving the quality of this
   manuscript.  This work was partly supported by ERC grant LEAP (no. 336845)
   and CIFAR Learning in Machines \& Brains program.
\end{document}